\documentclass[twocolumn, switch]{article} 

\usepackage{preprint}

\usepackage{amsmath, amsthm, amssymb, amsfonts}

\usepackage[numbers,square]{natbib}
\usepackage{natbib}

\usepackage{amsmath, amsfonts, bm, tabstackengine}

\def\ve{{\bm{e}}}

\def\vm{{\bm{m}}}

\def\vu{{\bm{u}}}
\def\vv{{\bm{v}}}

\def\vx{{\bm{x}}}
\def\vy{{\bm{y}}}

\def\mA{{\bm{A}}}

\def\mM{{\bm{M}}}

\def\mU{{\bm{U}}}
\def\mV{{\bm{V}}}

\def\mX{{\bm{X}}}

\usepackage{tikz}
\def\pres{{\color{green!50!black}\tikz\fill[scale=0.4](0,.35) -- (.25,0) -- (1,.7) -- (.25,.15) -- cycle;}}
\newcommand{\abse}{{\color{red!50!black}$\times$}}

\usepackage[utf8]{inputenc} 
\usepackage[T1]{fontenc}    
\usepackage{xcolor}		
\usepackage[colorlinks = true,
            linkcolor = purple,
            urlcolor  = blue,
            citecolor = cyan,
            anchorcolor = black]{hyperref}	
\usepackage{url}            
\usepackage{booktabs}       
\usepackage{amsfonts}       
\usepackage{nicefrac}       
\usepackage{microtype}      
\usepackage{xcolor}         
\usepackage{graphicx}
\usepackage{subcaption}
\usepackage{multirow}

\usepackage{amsmath}
\usepackage{bbm}
\usepackage{booktabs}
\usepackage{multirow}
\usepackage{siunitx}

\usepackage{color,soul}
\usepackage{wrapfig}
\usepackage{enumitem}

\usepackage{titlesec}
\titlespacing\section{0pt}{12pt plus 3pt minus 3pt}{1pt plus 1pt minus 1pt}
\titlespacing\subsection{0pt}{10pt plus 3pt minus 3pt}{1pt plus 1pt minus 1pt}
\titlespacing\subsubsection{0pt}{8pt plus 3pt minus 3pt}{1pt plus 1pt minus 1pt}

\title{PeakWeather: MeteoSwiss Weather Station Measurements\\for Spatiotemporal Deep Learning}

\usepackage{authblk}

\author[1,*]{Daniele Zambon}
\author[2,3,*]{Michele Cattaneo}
\author[1]{Ivan Marisca}
\author[2]{Jonas Bhend}
\author[2]{Daniele Nerini}
\author[1,4]{Cesare Alippi}

\affil[1]{Università della Svizzera italiana, IDSIA, Lugano, Switzerland.}
\affil[2]{Federal Office of Meteorology and Climatology MeteoSwiss, Zurich, Switzerland.}
\affil[3]{ETH Zürich and EPFL, Swiss Data Science Center, Switzerland.}
\affil[4]{Politecnico di Milano, Milan, Italy.}
\affil[*]{\small Equal contribution.}

\begin{document}

\twocolumn[ 
  \begin{@twocolumnfalse} 
  
\maketitle

\begin{abstract}
Accurate weather forecasts are essential for supporting a wide range of activities and decision-making processes, as well as mitigating the impacts of adverse weather events. While traditional numerical weather prediction (NWP) remains the cornerstone of operational forecasting, machine learning is emerging as a powerful alternative for fast, flexible, and scalable predictions. We introduce PeakWeather, a high-quality dataset of surface weather observations collected every 10 minutes over more than 8 years from the ground stations of the Federal Office of Meteorology and Climatology MeteoSwiss's measurement network. The dataset includes a diverse set of meteorological variables from 302 station locations distributed across Switzerland's complex topography and is complemented with topographical indices derived from digital height models for context. Ensemble forecasts from the currently operational high-resolution NWP model are provided as a baseline forecast against which to evaluate new approaches. The dataset's richness supports a broad spectrum of spatiotemporal tasks, including time series forecasting at various scales, graph structure learning, imputation, and virtual sensing. As such, PeakWeather serves as a real-world benchmark to advance both foundational machine learning research, meteorology, and sensor-based applications.
\end{abstract}
\vspace{0.35cm}

  \end{@twocolumnfalse} 
] 

\section{Introduction}

Weather forecasts provide essential information for protecting lives and properties, as well as supporting everyday decision-making.
The quality of these forecasts has steadily improved over recent decades, driven by advances in numerical weather prediction (NWP) and high-performance computing (HPC) \citep{bauer2015quiet}. NWP involves using mathematical models that simulate the weather based on current conditions of the atmosphere, land surfaces, and oceanic conditions. 
Both deterministic forecasts and ensembles of simulations are produced, with ensemble forecasts offering a probabilistic view of future states and serving as a critical tool for risk-informed decision-making. 
However, the complexity of the models, combined with the demand for high-resolution predictions in space and time, makes state-of-the-art NWP computationally expensive, often requiring several hours of runtime on modern HPC systems.

In parallel, the rise of deep learning (DL) has brought remarkable progress to a range of fields, including meteorology \citep{lam2023learning,bi2023accurate,pathak2022fourcastnet,price2025probabilistic}. 
Beyond breakthroughs in model architectures, this progress has been enabled by the availability of long-term, high-quality reanalysis datasets. Reanalyses combine modern NWP models with historical observational data, from ground stations, satellites, radar, ships, and aircraft, to produce a consistent and comprehensive estimate of past atmospheric states. DL models can learn to emulate the underlying physics captured in these massive datasets, enabling accurate and scalable data-driven weather forecasting. One of the key advantages of such models is their efficiency: once trained, they are significantly cheaper to run operationally than classical physics-based NWP. Nonetheless, these models still rely on NWP-generated analyses at inference time to initialize forecasts.

To overcome the need for NWP-based analysis and reanalysis data altogether, Direct Observation Prediction (DOP) methods have been proposed \citep{mcnally2024data}. 
DOP approaches aim to learn directly from raw observational data, eliminating the need for NWP data during both training and inference. When successfully implemented, DOP models can offer faster, more cost-effective forecasting pipelines and avoid potential biases introduced by physics-based simulations. DOP has shown promising results in short-range weather forecasting, particularly for lead times ranging from minutes to a few hours \citep{ravuri2021skilful,zhang2023skilful}, a domain known as nowcasting. A typical nowcasting application is the extrapolation of radar image sequences to predict imminent precipitation. While DOP has proven effective in short time horizons, NWP systems generally maintain an edge in predictive accuracy beyond a few hours.

\begin{figure}
  \centering
      \centering
      \includegraphics[width=\columnwidth]{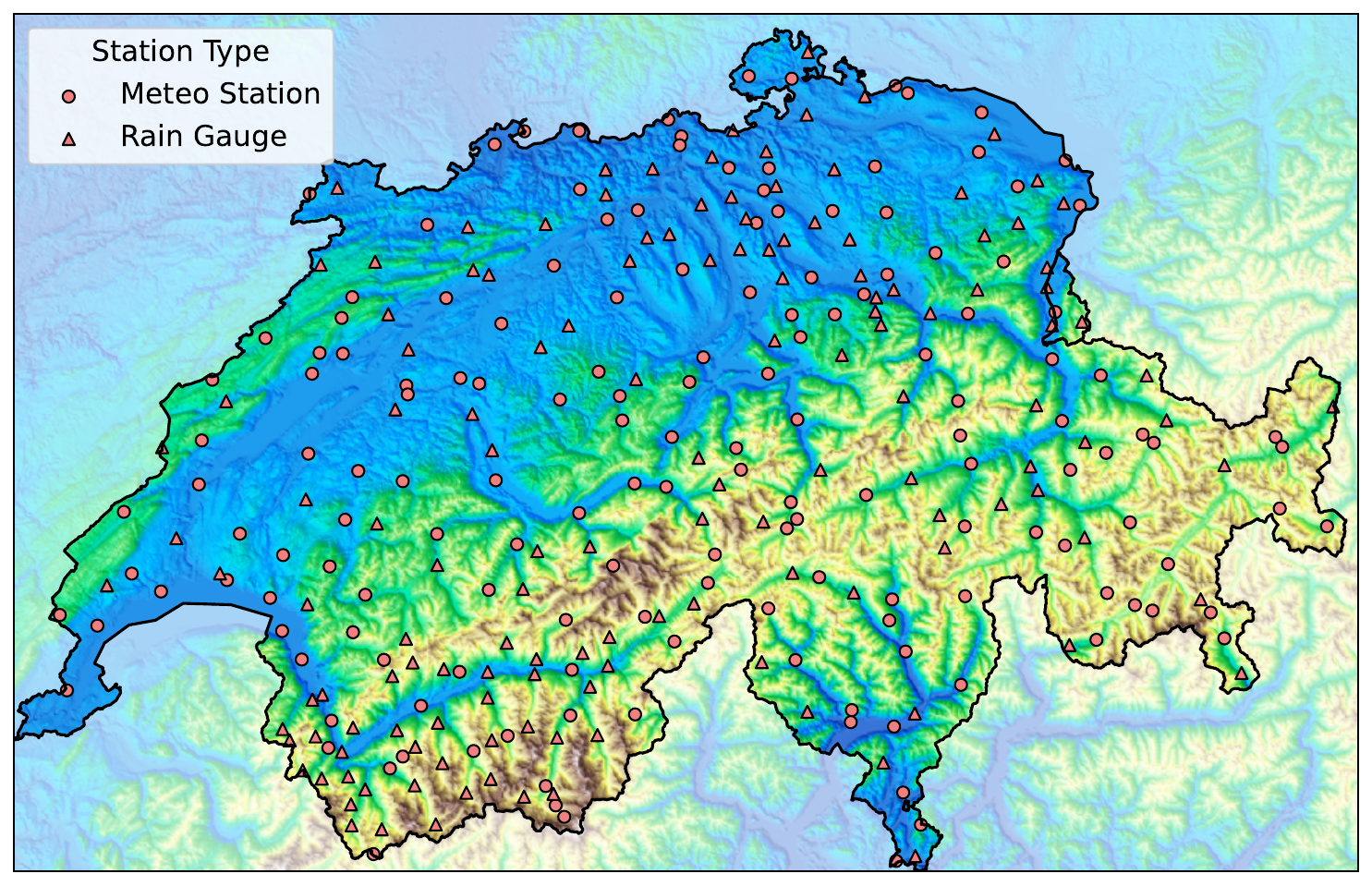}
  \caption{PeakWeather station locations across the Swiss territory.}
  \label{fig:stations-first-page}
\end{figure}

We introduce \textbf{PeakWeather}, a benchmarking dataset designed to advance deep learning research on spatiotemporal modeling. The dataset consists of high-resolution, ground-based meteorological measurements collected from SwissMetNet~\citep{SwissMetNet}, the network of automatic weather stations operated by the Swiss Federal Office of Meteorology and Climatology (MeteoSwiss). 
It includes eight key weather variables related to temperature, humidity, precipitation, pressure, sunshine, and wind, recorded every 10 minutes over a period of more than eight years. These observations come from 302 meticulously maintained and quality-validated stations \citep{SwissMetNetCert}, distributed throughout Switzerland, a country characterized by complex topography, shaped by the Swiss Alps, the Central Plateau, and the Jura Mountains; see also Figure~\ref{fig:stations-first-page}. In addition to meteorological variables, PeakWeather includes topographic descriptors derived from digital elevation models, capturing the unique characteristics of each station's surrounding area. 

To the best of our knowledge, this is the first openly available dataset that combines long-term, high-frequency ground station data with topographic context and state-of-the-art NWP baselines in complex terrain. 
The complexity of Switzerland's topography, combined with seasonalities and time variance of the meteorological quantities, poses unique challenges for modeling spatial and temporal dependencies. These factors make PeakWeather an ideal testbed for advancing data-driven methods in a wide array of machine learning tasks, including---but not limited to---time series forecasting, virtual sensing, graph structure learning, and missing data imputation. 
To support benchmarking, the dataset also includes ensemble forecasts from a state-of-the-art, high-resolution NWP model that has been operational at MeteoSwiss since May 2024. These forecasts serve as a strong physics-based reference point, enabling direct comparisons with machine learning approaches and promoting research in hybrid and alternative forecasting paradigms.

Our contributions are summarized as follows:
\begin{itemize}[leftmargin=2em]

\item 
Release of PeakWeather dataset \citep{PeakWeatherDataset}. The dataset contains: (i) A collection of high-quality validated multivariate time series from 302 weather stations comprising eight meteorological variables with a temporal resolution of 10 minutes acquired from January 1st, 2017 to October 13th, 2025; (ii) Topographical descriptors derived from a digital elevation model describing the Swiss terrain over a 50-meter grid; (iii) Numerical weather predictions from a state-of-the-art high-resolution NWP model. 

\item 
A framework-agnostic, easy-to-use and well-documented Python library \citep{PeakWeatherLibraryBaseCode}
for seamless interaction with the dataset. The library enables users to easily load, align, and preprocess the data, providing them in ready-to-use tensor formats suitable for deep learning workflows.

\item 
Evaluation of predictive models for wind forecasting---a task of high practical importance in domains such as renewable energy, aviation, and severe weather management. 
A range of deep learning architectures, including recurrent neural networks, spatiotemporal graph neural networks, and foundation models, are contrasted against forecasts from a state-of-the-art NWP model. 
Experiments extend to inductive temperature forecasting and study relevant modeling aspects, such as the use of graph-based modeling and station-specific components.
Reported results establish valuable baselines for future research and demonstrate the promising potential of deep learning models as viable alternatives to NWP models in specific weather prediction tasks.

\end{itemize}

The PeakWeather dataset is released under the CC-BY-4.0 license, with the associated library distributed under the BSD-3-Clause. 
With PeakWeather, we pursue a twofold objective: first, to facilitate access to high-quality meteorological data from MeteoSwiss, which is being progressively released as Open Government Data starting in 2025 \citep[OGD,][]{SwissMetNetDocumentation}; and second, to enable reproducible experimentation that advances foundational machine learning research, meteorology, and sensor-based applications.

\section{Related work}

One major category of meteorological datasets consists of gridded reanalysis data produced by NWP models, which assimilate past observations to estimate the weather state. 
One of the main driving forces behind the recent success of data-driven models is ECMWF Reanalysis v5 \citep[ERA5,][]{ERA5}. This dataset contains a detailed estimate of the global state of the atmosphere, land surface and oceans since the 1950s, based on the reanalysis of ECMWF's Integrated Forecasting System (IFS). Derived from ERA5, WeatherBench~\citep{rasp2020weatherbench1} and its recent successor WeatherBench2~\citep{rasp2024weatherbench2} offer standardized benchmarking datasets and tools, to facilitate the evaluation and comparison of machine learning models for weather forecasting.
For the task of statistical postprocessing \citep{vannitsem2021postproc}, where the goal is to correct systematic errors in NWP forecasts, EUPPBench~\citep{demaeyer2023euppbench} provides a dataset with aligned IFS forecasts and direct observations for a variety of meteorological variables. So far, however, only coarse-grained NWP data is available in EUPPBench and temporal resolution is limited to 6 hours.

Unlike the datasets above, PeakWeather supports DOP, enabling the development of models that rely solely on past observations and static features,  making the final model completely independent from NWP models. 
While datasets such as WeatherReal~\citep{jin2024weatherreal}, Weather2k~\citep{zhu2023weather2k}, Monash Weather~\citep{godahewa2021monash}, and StationBench~\citep{stationbench} share this DOP focus, PeakWeather distinguishes itself through its unique combination of characteristics not available in other datasets: quality-controlled data with a high spatial density and 10-minute temporal resolution, spanning over eight years across Switzerland's challenging topography, and physics-based forecasts from an operational NWP model for comparison.

\section{PeakWeather dataset}\label{sec:dataset}

We introduce PeakWeather, a publicly available dataset hosted on Hugging Face \citep{PeakWeatherDataset}. The dataset is organized into three core components: (i) high-resolution historical surface observations, (ii) static topographic features derived from elevation data, and (iii) ensemble forecasts from an operational NWP model. To facilitate easy access and streamlined preprocessing, we also release an open-source Python library \citep{PeakWeatherLibraryBaseCode} tailored to work seamlessly with the dataset.

\paragraph{Meteorological observations at stations}
Meteorological observations come from the SwissMetNet network, Switzerland's reference ground-based measurement system for weather and climate monitoring operated by MeteoSwiss. 
The network in PeakWeather comprises 302 stations in total: 160 standard meteorological stations measuring various weather and climate parameters, complemented by 142 additional stations measuring precipitation. 
Stations are labeled as either meteorological stations or rain gauges, as shown in Figure~\ref{fig:stations-first-page}. 
The average distance to the 5 nearest neighbors is 18km across the 160 core meteorological stations, and 12km when all stations are included.
The stations follow the World Meteorological Organization (WMO) standards, use consistent sensors, and are regularly maintained \citep{SwissMetNetCert}. The measurement data undergo automatic and manual quality controls. 
From the range of weather parameters available, we select air temperature, relative humidity, atmospheric pressure, sunshine duration, wind speed and direction, wind gusts, and the amount of precipitation, as summarized in Table \ref{tab:meteo_variables}; the table also reports the unique identifier (Short name) consistent with the SwissMetNet naming convention and Open Government Data (OGD) program \citep{SwissMetNetDocumentation}. Statistics about parameters measured by different stations are reported in Table~\ref{tab:missing_values}, highlighting variability across stations. 

\paragraph{Topographic descriptors}
Features of the terrain play a crucial role in weather and climate processes across multiple scales. By integrating detailed terrain information, the prediction accuracy of data-driven models can be enhanced as demonstrated in Section~\ref{sec:experiments}.
To this end, PeakWeather includes a high-resolution digital height model (DHM) with a spatial resolution of $50$ meters, capturing the bare ground elevation without vegetation and buildings. The DHM is derived from the DHM25 \citep{swisstopoDHM25} product by the Swiss Federal Office of Topography swisstopo; a visualization is provided in Figure~\ref{fig:stations-first-page}.
In addition, we provide various derived topographical features that represent physical characteristics of the terrain and its surroundings. 
These features%
    \footnote{The implementation of these features has been open-sourced by MeteoSwiss \url{https://github.com/MeteoSwiss/topo-descriptors}}
are computed considering neighborhoods of $2$ and $10$ kilometers to cover phenomena at various scales and include the topographic position index (TPI) which describes whether a point is above or below the average elevation of its surroundings, the standard deviation (STD) of the elevation, the terrain aspect, the terrain slope, the west-east and south-north gradients of the terrain; see also Appendix~\ref{app:viz-topo}.

\paragraph{Numerical weather predictions}  
We include forecasts from ICON-CH1-EPS, one of the NWP models operational at MeteoSwiss since May 2024, as a strong physics-based baseline. ICON-CH1-EPS is a regional, high-resolution, and ensemble setup of the ICOsahedral Non-hydrostatic \citep[ICON,][]{zangl2015icon} model for Switzerland. It runs every 3 hours and produces an 11-member ensemble: one control run and ten perturbed members. Operating at $\sim$1km grid resolution, it generates forecasts up to 33 hours and the analysis step (lead time 0). PeakWeather includes hourly forecasts of the same variables and extracted at each station's nearest model grid cell starting from May 14th, 2024. Refer to Appendix~\ref{app:icon} for additional details.

\begin{table*}
\centering\caption{Meteorological variables present in the dataset.}
\resizebox{.95\textwidth}{!}{%
\begin{tabular}{rclcc}
\toprule
\textbf{Variable name} & \textbf{Short name} & \textbf{Description} & \textbf{Aggregation} & \textbf{Unit} \\ 
\midrule
\texttt{temperature} & \texttt{tre200s0} & Air temperature 2m above the ground & Current value & Celsius \\ 
\texttt{humidity} & \texttt{ure200s0} & Relative air humidity 2m above the ground & Current value & \% \\  
\texttt{precipitation} & \texttt{rre150z0} & Precipitation & 10min total & mm \\ 
\texttt{sunshine} & \texttt{sre000z0} & Sunshine duration & 10min total & min \\ 
\texttt{pressure} & \texttt{prestas0} & Atmospheric pressure at barometric altitude (QFE) & Current value & hPa \\  
\texttt{wind\_speed} & \texttt{fkl010z0} & Wind speed & 10min mean & m/s \\ 
\texttt{wind\_gust} & \texttt{fkl010z1} & Wind gust peak of 1 second & 10min max & m/s \\
\texttt{wind\_direction} & \texttt{dkl010z0} & Wind from direction, measured clockwise from north. & 10min circ.~mean & degree \\  
\bottomrule
\end{tabular}
}
\label{tab:meteo_variables}
\end{table*}

\begin{table*}
\centering
\caption{Number of stations per variable and station type, shown as ``\# in Jan 2017 / \# in Oct 2025.'' Missing data percentages are computed only for stations that recorded the variable.}
\resizebox{.9\textwidth}{!}{%
\begin{tabular}{rcccccccc}
\toprule
& \textbf{Temp.} & \textbf{Humidity} & \textbf{Precip.} & \textbf{Sunshine} & \textbf{Pressure} & \textbf{Wind speed} & \textbf{Wind gusts} & \textbf{Wind dir.} \\
\midrule
Rain gauges    & 41/38 & 0/0 & 124/141 & 0/0  & 0/0  & 0/0  & 0/0  & 0/0  \\
Meteo stations & 148/148 & 148/148 & 138/138& 103/129 & 137/138 & 149/150 & 150/150 & 150/150 \\
Missing values & 1.94\% & 0.96\% & 3.51\% & 13.02\% & 1.39\% & 1.46\% & 1.08\% & 1.10\% \\
\bottomrule
\end{tabular}
}
\label{tab:missing_values}
\end{table*}

\paragraph{Practical considerations for data use} One potential source of inconsistencies is sensor or station relocation, which---even in a carefully managed network like SwissMetNet---is sometimes necessary. 
To ensure transparency, PeakWeather also includes detailed metadata on all sensor relocations which, however, are generally minor: 95\% of displacements are under 110 meters, with an average of 97 meters. Although the stations and data are quality controlled, it should be noted that artifacts may still be present (e.g., due to frozen anemometers), and that the temperature sensors at rain gauges can be of inferior quality, as they are intended to discriminate between solid and liquid precipitation. 

\paragraph{Missing data}
Missing data in PeakWeather stems from both temporary measurement gaps and the fact that not all 302 SwissMetNet stations are equipped with the same set of sensors. Sensor configurations also evolve due to operational requirements. To give an accurate overview, Table~\ref{tab:missing_values} presents the number of stations recording each variable at the start and end of the dataset (January 2017 and October 2025), along with missing value statistics computed only for stations that have measured each variable at some point. Over the entire period, the overall proportion of missing data is approximately 3\%. Figure~\ref{fig:missing_values} in the appendix shows the sensor availability over time. Missing values are represented as NaNs and can be handled using built-in utilities (e.g., forward fill or zero-fill). A binary availability mask is also provided by PeakWeather, allowing users to easily identify and handle invalid observations at each time step, station, and variable level.

\section{Machine learning tasks on PeakWeather}\label{sec:tasks}

The PeakWeather dataset supports a wide range of machine learning tasks. 
While forecasting is likely the most common application, PeakWeather also enables the investigation of additional tasks and modeling aspects arising from the dataset's distinctive characteristics.
In this section, we outline a set of representative examples that, while not exhaustive, illustrate the dataset's versatility for both foundational research and practical applications in data-driven weather modeling.

\paragraph{Notation}
We formalize the dataset as consisting of $N$ multivariate time series, each collected from a different station. Let $\vx_t^i \in \mathbb{R}^{D_x}$ denote the $D_x$-dimensional observation at time step $t$ from station $i \in \{1, \dots, N\}$;
here, $N$ denotes the total number of weather stations in the dataset and $D_x$ the number of possible measured variables (channels) across all stations.
Regular and synchronous sampling is assumed. However, as mentioned above, not all stations measure the same weather quantities, and not all sensors are available at all time steps (see Table~\ref{tab:missing_values}). We account for such data (un)availability by appropriately padding the time series, thereby maintaining a tabular representation. Accordingly, an auxiliary binary mask $\vm^i_t \in \{0, 1\}^{D_x}$ is introduced at each time step $t$ and station $i$ to flag the data availability at the level of the stations, channels, and time steps. In particular, we set $\vm_t^i[d]=1$ if the $d$-th of the $D_x$ channels of the $i$-th station at time step $t$ is valid, and $\vm_t^i[d]=0$ otherwise.
For brevity, denote by $\mX_{t}\in\mathbb R^{N\times D_x}$ the stack of all observations $\{\vx_t^i\}_{i=1}^N$ at time step $t$, by $\mX_{t:t+T}\in\mathbb R^{T\times N\times D_x}$ the observations from time steps in $\{t,t+1,\dots t+T-1\}$ and by $\mX_{<t}$ all observations up to time step $t$; similarly, $\mM_t\in\mathbb R^{N\times D_x}$ denotes the mask across all stations at time step $t$. The PeakWeather library provides functionalities to generate such a mask.

\paragraph{Data-generating process} 
As the time series are correlated with each other, we model the data-generating process as a time-invariant spatiotemporal stochastic process such that
\begin{equation}\label{eq:system-model}
    \vx_t^i \sim p^i(\vx_t^i | \mX_{<t}, \mU_{\le t}, \mV),  \qquad \forall i=1,\dots, N,
\end{equation}
where $\mV\in \mathbb R^{N\times d_v}$ and $\mU_t \in \mathbb R^{N \times d_u}$ are exogenous variables describing additional information related to the measured variables $\vx_t^i$, such as station-specific characteristics, seasonalities, as well as the topographic descriptors from the DHM (see Section~\ref{sec:dataset}).
Note that the process of Equation~\ref{eq:system-model} generating all time series is not necessarily the same, i.e., $p^i \ne p^j$ if $i \ne j$, while the assumption of time invariance remains valid.

\paragraph{Time series forecasting} 
This task consists of predicting future values of meteorological variables of interest (e.g., temperature, wind speed) at individual stations, possibly all, by using historical observations and, whenever available, exogenous variables affecting the system. 
Denote by $\vy_t^i\in\mathbb R^{D_y}$ the vector of target variables and formulate the task as that of learning a model $p_\theta$ to predict $\vy^i_{t+h}$  as 
\begin{multline}
\hat\vy_{t+h}^i\sim p_\theta(\vy_{t+h}^i| \mX_{t-W:t}, \mM_{t-W:t}, \mU_{t-W:t+h+1}, \mV) 
\\\approx
p^i(\vy_{t+h}^i | \mX_{<t}, \mU_{\le t+h}, \mV),
\end{multline}
for a set of lead times corresponding to $h=0,\dots,H-1$ and stations $i$ of interest; $W\ge 1$ denotes the time window length used to make the predictions. 
This task can be addressed at multiple temporal resolutions and forecast horizons, and can leverage both spatial and temporal dependencies to predict weather conditions across multiple stations simultaneously. 

Wind forecasting is the main application considered in the experiments in Section~\ref{sec:experiments}. Accurate wind forecasts provide useful information for diverse applications such as severe weather and renewable energy management. Forecasting wind in complex terrain, however, is challenging due to the dominant influence of local conditions. Exposure to prevailing large-scale winds and related effects such as channeling and blocking affect wind speeds at individual locations. In addition, distinct phenomena such as mountain valley wind systems dominate the temporal evolution. Temperature forecasting is studied in Appendix~\ref{app:temperature-extra}.

\paragraph{Local effects and node-specific components}
As anticipated in \eqref{eq:system-model}, time series observed at different locations may exhibit {local effects}, namely variations that are specific to individual locations or small regions. Such specificity can arise from, e.g., unobserved factors and station-specific characteristics. In the case of PeakWeather, the complex Swiss terrain in which measurement stations are deployed plays a role. 
Accounting for local effects poses a significant modeling challenge, as it requires balancing the extraction of global patterns common to all time series with the ability to specialize predictions to individual ones, under a less favorable sample size \citep{montero-manso2021principles}. 
In the experimental section below, we demonstrate the importance of incorporating station-specific components for achieving accurate predictions. In particular, we show the positive impact of conditioning predictions on station-level topographic descriptors $\vv^i$ as well as on learnable station-specific embeddings $\ve^i\in \mathbb R^{D_e}$.

\paragraph{Relational inductive biases for time series processing}
Beyond local components, time series processing can be significantly improved by exploiting the underlying (functional) dependencies among series acquired from spatially distributed stations. Rather than modeling each sequence in isolation, one can incorporate relational information as an inductive bias, allowing models to condition predictions on related time series and reduce the risk of overfitting to local, spurious patterns.
These dependencies can be encoded via a graph structure represented by an adjacency matrix $\mA \in \{0,1\}^{N \times N}$, where each node corresponds to a station and non-zero entries denote pairwise relationships. When needed, $\mA$ may be enriched with edge weights or attributes reflecting the strength or type of interaction (e.g., distance, elevation difference, empirical correlation). This setup is at the core of graph deep learning approaches for time series processing \citep{cini2023graph}, where message-passing operators like graph neural networks (GNN) are exploited to exchange information across the graph topology.
In the context of meteorological data, such relations are often assumed to be time-invariant. However, changes to the sensor network---such as the addition, removal, or relocation of stations---may require considering a dynamic or evolving graph. 
While spatial proximity naturally induces correlation (i.e., signal smoothness), this assumption breaks down at larger scales or in complex terrains. In alpine regions, for instance, valleys, ridges, and altitude gradients introduce significant variability, making long-range dependencies non-trivial to infer. Particularly in datasets like PeakWeather, this motivates considering relations learned in a data-driven manner and tailored to the given task, as discussed in the next paragraph. 

\paragraph{Graph Structure Learning}  
Better predictions can be obtained by leveraging relational information among the time series, as demonstrated in the next Section~\ref{sec:experiments}. However, it is not always evident which stations/variables would benefit from exchanging information. While it is common to extract a graph of binary relations from, e.g., physical proximity, correlations among time series, graph structure learning has emerged as a task where relations can be treated as latent variables that are learned in a data-driven manner, often optimizing a prediction task of interest. Given the complex Swiss terrain and the specificity of each station's location, performing graph structure learning in PeakWeather appears sound and insightful.

\paragraph{Missing data imputation and virtual sensing} 
Another challenge posed by real-world data concerns temporal gaps in the measurements and unavailable observations at certain locations.
A related task is broadly known as spatial interpolation. It aims to estimate unmeasured weather variables by leveraging spatial patterns learned from nearby stations, topographical features, and, when available, related variables at the same location. 
While missing data imputation focuses on filling temporal gaps (missing or erroneous data), virtual sensing refers to predicting variables at locations without any prior measurements. 
Notably, addressing these tasks may involve conditioning predictions on observations at future time steps. 
Despite the high reliability of SwissMetNet, missing observations remain inevitable (Table~\ref{tab:missing_values}). Moreover, the continuous updates to the SwissMetNet network, where new stations are added, some are reconfigured, and others are removed, motivate considering inductive learning methods, where model predictions can extend to new, unseen stations based on their spatial and contextual characteristics.
In Appendix~\ref{app:temperature-extra}, we experiment on this inductive setting considering the temperature as target variable.

\section{Experiments: a use case in wind forecasting}\label{sec:experiments}

In this section, we consider wind forecasting as a representative spatiotemporal modeling task to showcase the potential and breadth of PeakWeather.
We evaluate the capabilities and advantages of deep spatiotemporal models in comparison to the reference NWP model based on atmospheric simulations.
Alongside forecasting, we demonstrate the versatility of PeakWeather to study different related learning challenges, including the modeling of station-specific effects and the learning of relational information directly from data. In particular, we show the positive impact of incorporating topographical descriptors, learnable station-specific embeddings, and graph structure learning.

\paragraph{Wind forecasting setup}
We consider the task of predicting wind direction and speed for the next 24 hours. Specifically, we consider hourly data and set the wind horizontal velocity vector as target $\vy_t^i\in\mathbb R^2$. The components of $\vy_t^i$, also called $u$ and $v$ components, are constructed from variables \texttt{wind\_speed} and \texttt{wind\_direction} and indicate the eastward and northward wind components. For this task, we consider data from the 160 meteorological stations and with static attributes encoding the latitude, longitude, station height, and other topographic features.
Notably, resampling from 10-minute resolution to 1-hour (while accounting for variable-specific aggregation strategies), $u$ and $v$ component extraction, and weather station filtering can all be carried out via the utilities in the PeakWeather dataset library. 
The considered models described in the following sections are trained on the historical data before October 2024 (i.e., almost eight years), tested on the last year of available data, and validated on a held-out set of training data from October 2023 to September 2024.
A graph similar to that in Figure~\ref{fig:stations_graph} is constructed by connecting stations based on their horizontal Euclidean distance; the graph is passed as input to all models requiring one.
Further details on the training procedure, model selection, and graph construction are provided in Appendix~\ref{app:exp-details}.

\paragraph{Evaluation methodology}
We train models minimizing the Energy Score (ES) \citep{gneiting2007strictly}---an extension of the Continuous Ranked Probability Score (CRPS) for multivariate predictions---which evaluates the quality of probabilistic forecasts accounting for both the calibration and sharpness of predicted distributions; the ES is estimated and optimized via Monte Carlo sampling. Models are then evaluated on the test set using both probabilistic and point prediction metrics. Specifically, we report the ES for the bivariate wind velocity, and the mean absolute error on the same vector (MAE Velocity) to assess point-forecast accuracy. In addition, we include the MAE for the wind speed (MAE Speed) and for the wind direction (MAE Direction), the latter computed as the mean angular distance between the target and predicted wind velocity. All metrics are reported across several lead times up to 24 hours. The metrics are evaluated considering 100 Monte Carlo samples (11 samples for ICON), with the MAEs based on the sample median. Missing observations are appropriately masked during evaluation.

\subsection{Forecasting models and baselines} 

We consider two main families of deep learning forecasting architectures, \emph{temporal models} like recurrent neural networks and \emph{spatiotemporal models} based on message passing and graph neural networks~\citep{jin2024survey}.

The temporal models \textbf{RNN} and \textbf{TCN} are multilayer networks processing each station independently from the others. 
The architectures feature an encoder block to aggregate the input $\vx_{t-w}^i$ and exogenous variables $\vu_{t-w}^i$ and $\vv^i$, before feeding them to the RNN/TCN temporal processing, and a decoder responsible for making forecasts at the desired lead times; both encoder and decoder are implemented as multilayer perceptrons. 
The spatiotemporal models are \textbf{MP-RNN} and \textbf{MP-TCN} and implement a time-then-space architecture \citep{cini2023graph}, following the same high-level architecture of the temporal models, but composing GNN layers on top of the temporal processing, before issuing predictions via the decoder.

As competitive models from the literature, we consider spatiotemporal models \textbf{DCRNN} \citep{li2018diffusion} and \textbf{AGCRN} \citep{bai2020adaptive}, based on RNNs, and \textbf{GraphWaveNet} \citep{wu2019graph}, based on temporal convolution. As temporal models, we include the foundation model \textbf{Chronos-2} \citep{ansari2025chronos}, to reflect the growing relevance of large pretrained models in time series forecasting. 

For reference, two versions of the persistence model (PM) are implemented: one particularly relevant for short-term, \textbf{PM-st}, and the other, \textbf{PM-day}, intended to replicate the daily patterns of the target variables. PM-st is a baseline model that predicts $\hat\vy_{t+h}^i=\vy_{t-1}^i$ as the last available observation of the target for each lead time $h=0,\dots,H-1$, a valid assumption for very short-term forecasts. Conversely, the output of PM-day is $\hat\vy_{t+h}^i=\vy_{t+h-24}^i$, recalling that in this specific problem instance we are considering hourly data and a forecasting horizon of $H=24$. While less evident than with other weather quantities, the wind can have a diurnal seasonality, especially in mountainous valleys. 

Finally, the \textbf{ICON} numerical predictions (Section~\ref{sec:dataset}) provided within PeakWeather and obtained from ICON-CH1-EPS are compared with the above models.

\subsection{Results}\label{sec:results}

We present four sets of experiments to illustrate how PeakWeather can be used in practice and to highlight its value for the deep learning and spatiotemporal modeling research communities. 
\textbf{(i)} We investigate the impact of incorporating station-specific information---a core feature of PeakWeather---highlighting the presence of local effects in the data. 
\textbf{(ii)} We evaluate the benefits of graph-based models and of learning relational information directly from the data. 
\textbf{(iii)} We benchmark the wind prediction accuracy of a range of models, showcasing the competitiveness of graph deep learning approaches for spatiotemporal modeling, and providing reference prediction accuracies for future research. 
Finally, \textbf{(iv)} we compare the best-performing models against forecasts from the NWP model ICON, demonstrating the challenging nature of the task and highlighting the potential held by deep learning methods in this field.

\begin{table}
\caption{Assessment of the impact on the prediction performance of including station-specific features and learning a graph structure from data. The family of models based on temporal convolution is considered here; for other models see Table~\ref{tab:local-effects-gsl-extended}. The first columns indicate whether the corresponding model relies on station-specific topographic descriptors $\vv^i$~(Topo.), learnable node embeddings $\ve^i$~(Emb.), an input graph (In.G.), and a graph structure learned from the data (GSL). Results are aggregated over five runs with different random seeds and are reported as mean $\pm$ standard deviation. For both metrics, the lower the better. Symbol ``-'' indicates non-applicability. 
}
\label{tab:local-effects}
\label{tab:gsl}
\centering
\resizebox{\columnwidth}{!}{%
\begin{tabular}{ccccccc}
\multicolumn{7}{c}{\emph{TCN Family}}\\
\toprule
 & \multicolumn{4}{c}{Model Components} &  \multicolumn{2}{c}{Metrics} \\
 & Topo. & Emb. & In.G. & {GSL} & MAE Vel. & En. Score \\
 \midrule
\multirow{4}{*}{\rotatebox{90}{Temporal}}
 & \abse & \abse  & - & - & 1.112\textsubscript{{$\pm$}0.002} & 1.262\textsubscript{{$\pm$}0.002} \\
 & \pres & \abse  & - & - & 1.077\textsubscript{{$\pm$}0.003} & 1.215\textsubscript{{$\pm$}0.004} \\
 & \abse & \pres  & - & - & 1.076\textsubscript{{$\pm$}0.001} & 1.213\textsubscript{{$\pm$}0.001} \\
 & \pres & \pres  & - & - & 1.076\textsubscript{{$\pm$}0.002} & 1.212\textsubscript{{$\pm$}0.002} \\
 \midrule
\multirow{6}{*}{\rotatebox{90}{Spatiotemporal}}
 & \abse & \abse & \pres & \abse & 1.024\textsubscript{{$\pm$}0.004} & 1.154\textsubscript{{$\pm$}0.005}\\
 & \pres & \abse & \pres & \abse & 1.014\textsubscript{{$\pm$}0.003} & 1.141\textsubscript{{$\pm$}0.003}\\
 & \abse & \pres & \pres & \abse & 1.011\textsubscript{{$\pm$}0.001} & 1.137\textsubscript{{$\pm$}0.002}\\
 & \pres & \pres & \pres & \abse & 1.010\textsubscript{{$\pm$}0.002} & 1.135\textsubscript{{$\pm$}0.002}\\
 & \pres & \pres & \abse & \pres & 1.001\textsubscript{{$\pm$}0.003} & 1.125\textsubscript{{$\pm$}0.004} \\
 & \pres & \pres & \pres & \pres & 0.999\textsubscript{{$\pm$}0.001} & 1.125\textsubscript{{$\pm$}0.002} \\
\bottomrule
\end{tabular}}
\end{table}

\paragraph{Relevance of station-specific features}
Table~\ref{tab:local-effects} reports test performance for temporal and spatiotemporal models trained with and without station-specific attributes, namely static station descriptors $\vv^i$ and node-level learnable embeddings $\ve^i\in \mathbb R^{D_e}$. Unlike the topographic features in $\vv^i$, the embeddings $\ve^i$ are learned end-to-end together with the other model parameters and then used to condition both the encoder and decoder.
For purely temporal models, incorporating either $\vv^i$, $\ve^i$, or both leads to substantial improvements in accuracy, highlighting the importance of station-level information. 
Spatiotemporal models also benefit from the inclusion of station-specific features, although the gain is less pronounced.
This suggests that message passing already provides a degree of localization by conditioning predictions on related time series, thereby reducing the contribution of station embeddings. 
Nevertheless, spatiotemporal models consistently outperform temporal ones overall.

\paragraph{Relevance of graph structure learning}
Table~\ref{tab:local-effects} further shows that replacing or augmenting the (heuristic) proximity-based graph with a data-driven graph yields significant improvements in prediction accuracy. 
This learned relational structure is jointly optimized with the remaining model parameters on the downstream task. 

Similar trends are observed across RNN, MP-RNN, and GraphWaveNet architectures that we report in Table~\ref{tab:local-effects-gsl-extended}. 
Overall, these results emphasize the importance of accounting for both station-specific features and relational information, and demonstrate that such structure can be effectively learned from data, tailored to the considered task.

\begin{figure*}
\centering
\includegraphics[width=.95\textwidth]{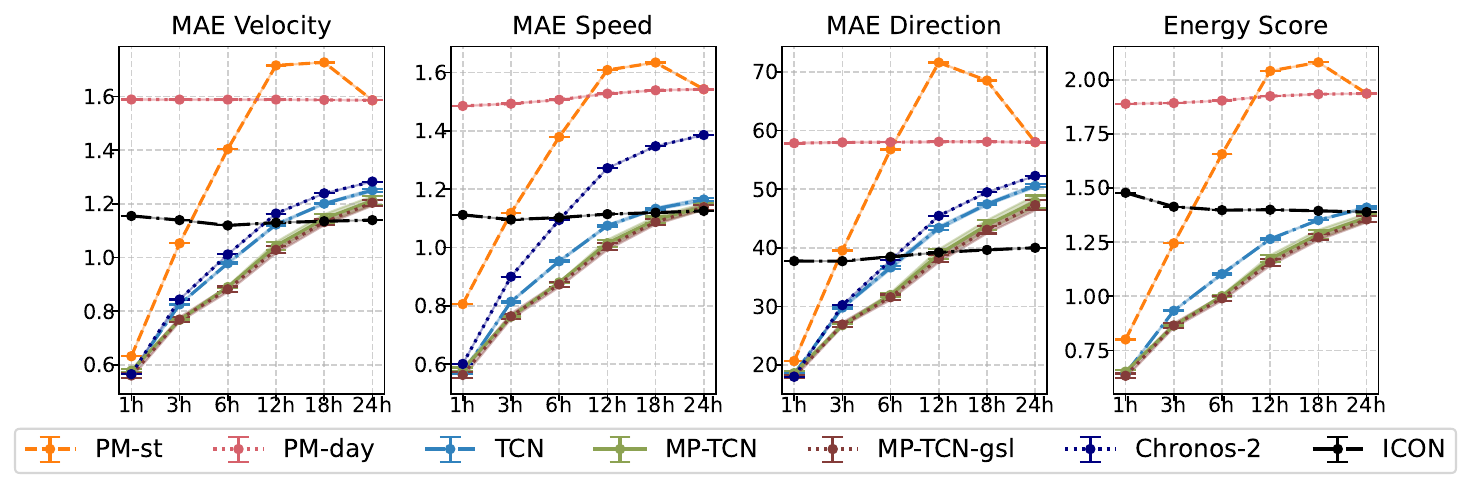}
\caption{Performance comparison of deep learning models with ICON and persistence model baselines for wind forecasting. Results are averaged over five runs with different random seeds, except for ICON and Chronos-2. Shaded areas indicate $\pm$3 standard deviations. The persistence models, ICON and Chronos-2 require no training. The variability of the persistence models and ICON arises solely from Monte Carlo sampling during evaluation and the ensemble forecasts, respectively. Chronos-2 provides quantile predictions of the velocity, which are used to compute the speed and direction.}
\label{fig:wind-1d}
\end{figure*}

\paragraph{Forecasting accuracy}
Figure~\ref{fig:wind-1d} shows forecasting performance across lead times for selected models. All trained models account for topographic attributes $\vv^i$ and learnable embeddings $\ve^i$. While MP-TCN operates on the proximity graph, {MP-TCN-gsl} replaces it with a learned one.
PM-st prediction accuracy provides reference performance, particularly relevant in the short term. As the horizon increases, PM-day marginally outperforms PM-st, although the difference is minor within the considered 24 hours; as expected, PM-st and PM-day coincide at 24h predictions.
All deep learning models substantially outperform the PM baselines across all horizons, both in terms of MAE and probabilistic accuracy, measured by the ES metric. Among them, the spatiotemporal models achieve the best overall performance, underscoring the value of incorporating spatial structure; these results are also confirmed by Table~\ref{tab:wind} in Appendix~\ref{app:wind-extra}.
As expected, performance degrades with increasing forecast horizon. This trend holds for all models, with the exception of the persistence baselines: PM-day effectively remains fixed at 24h forecasting horizon, while PM-st benefits from daily wind patterns. Notably, wind direction is consistently harder to predict than wind speed as evidenced by the smaller improvements over the PM baselines. Similar results abserved for temperature forecasting in Appendix~\ref{app:temperature-extra}.

\paragraph{Comparison with ICON}
We observe that ICON demonstrates relatively stable performance across all lead times, including short-term forecasts. However, its accuracy does not improve significantly at near-term horizons; this is likely due to spatial mismatches between the gridded NWP output and the fine-scale conditions at station locations and consequent challenges in assimilating such local information in the model. 
Comparing ICON to the deep learning models reveals that the MP-TCN deliver more accurate predictions in the nowcasting regime, and up to approximately 12 hours ahead. For longer horizons, and particularly for wind direction, ICON outperforms the data-driven approaches, suggesting that physics-based models still hold an advantage in capturing larger-scale dynamics over extended periods. Notably, the DL models are more competitive when assessed with the Energy Score, as also visible in Figure~\ref{fig:forecasts}. This is likely due to the fact that data-driven models better represent the local specifics at station measurement locations.

These findings align with our earlier discussion: deep learning models excel at short-term forecasts, offering efficient and accurate predictions, while NWP models currently maintain superior performance for longer horizons. This highlights an important frontier for machine learning research in spatiotemporal modeling.

\section{Conclusion}

The paper introduces PeakWeather, a high-resolution dataset of validated ground-based weather measurements collected from the SwissMetNet network operated by MeteoSwiss. Designed to support and accelerate research in spatiotemporal deep learning, the dataset provides dense and diverse weather observations from 302 stations distributed across Switzerland. The dataset includes eight meteorological variables sampled every 10 minutes for more than eight years. Notably, the data are enriched with topographical descriptors to account for the complex Swiss terrain and with ensemble forecasts from a state-of-the-art NWP model that is currently operational at MeteoSwiss for benchmarking.

Beyond its scale and quality, PeakWeather distinguishes itself by supporting a wide range of machine learning tasks. These include time series forecasting, virtual sensing (spatial interpolation), graph structure learning, and missing data imputation. PeakWeather data are provided alongside a framework-agnostic library that offers data in ready-to-use tensor formats for seamless integration into modern deep learning workflows.
Experiments on wind forecasting conclude the paper, showcasing the timeliness of PeakWeather, its relevant features for spatiotemporal modeling, and reinforcing the emerging role of deep learning as a complementary tool to traditional NWP systems. We are confident that PeakWeather will foster both fundamental machine learning research as well as advance data-driven approaches in meteorological modeling.

\section*{Acknowledgments}
This work was supported by the Swiss National Science Foundation project FNS 204061: \emph{HORD GNN: Higher-Order Relations and Dynamics in Graph Neural Networks}.

\bibliography{biblio}

\appendix 
\onecolumn

\section{Visualization of topographic descriptors and station distribution}\label{app:viz-topo}

\begin{figure}[b!]
  \centering
  \begin{subfigure}[t]{0.48\textwidth}
      \centering
      \includegraphics[width=\textwidth]{figures/stations.pdf}
      \caption{}
      \label{fig:stations}
  \end{subfigure}
  \begin{subfigure}[t]{0.48\textwidth}
      \centering
      \includegraphics[width=\textwidth]{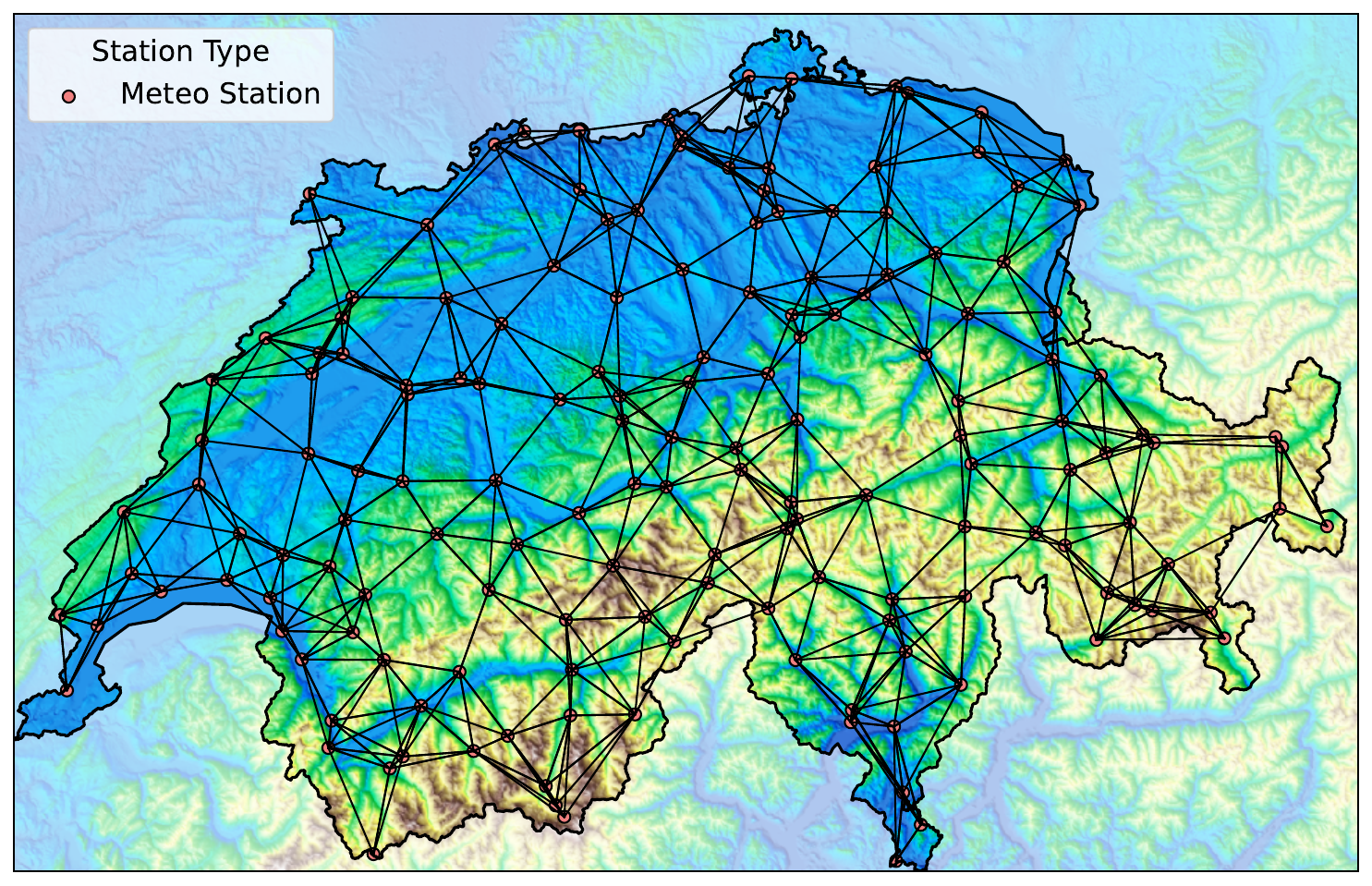}
      \caption{}
      \label{fig:stations_graph}
  \end{subfigure}
  \begin{subfigure}[t]{1.0\textwidth}
      \centering
      \includegraphics[width=\textwidth]{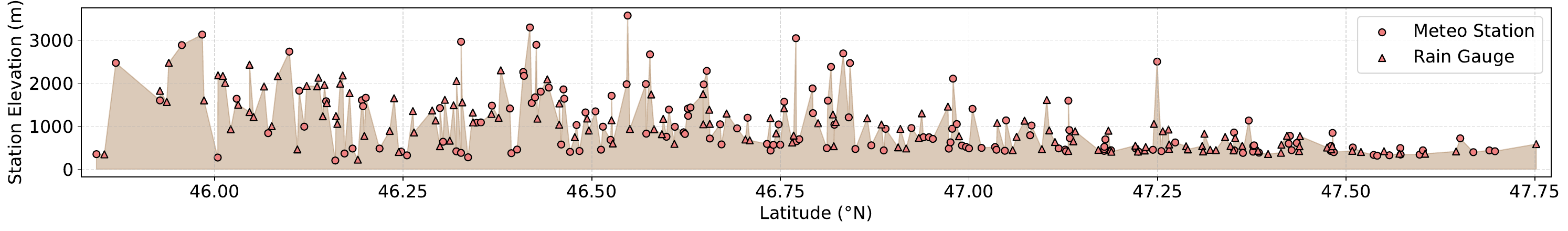}
      \includegraphics[width=\textwidth]{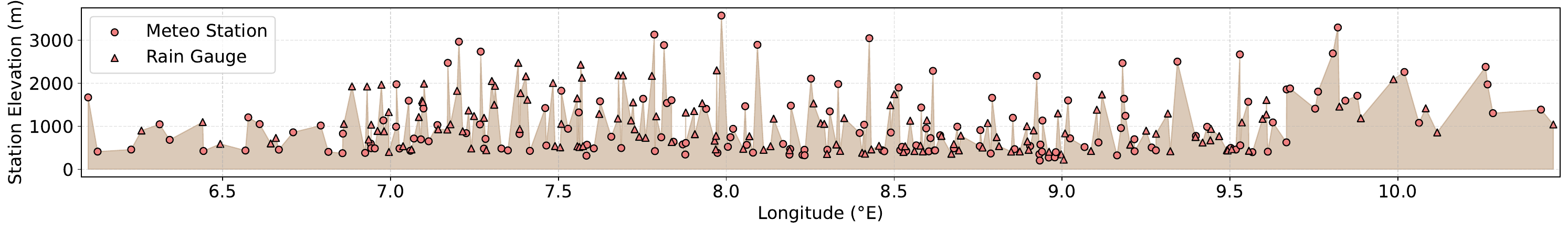}
      \caption{}
      \label{fig:height_cov_lat}
      \label{fig:height_cov_lon}
      \label{fig:height_cov_latlon}
  \end{subfigure}
  \caption{Visualizations of the placement of PeakWeather stations. Panel \ref{fig:stations}) distribution of the stations across the Swiss territory; circles denote meteorological stations, while triangles denote rain gauges. Panel \ref{fig:stations_graph}) a graph obtained using the geographical distance to compute a similarity heuristic for the meteorological stations. Panel \ref{fig:height_cov_latlon}) vertical coverage of the stations.}
  \label{fig:stations_and_graph}
\end{figure}

Figure \ref{fig:stations} shows the distribution of meteorological stations and rain gauges across the Swiss terrain. Figure \ref{fig:stations_graph} instead displays an example of a heuristically defined graph based on the distance between locations. 
Figure \ref{fig:height_cov_latlon} shows the station elevation across longitude and latitude.
Figure \ref{fig:topo_descriptors} shows the topographic features included in PeakWeather, which are computed at both 2 km and 10 km spatial scales. The Aspect (Figures \ref{fig:aspect10k} and \ref{fig:aspect2k}) is defined as the azimuth (in degrees from north) of the steepest downslope direction. The Standard Deviation (STD) (Figures \ref{fig:std10k} and \ref{fig:std2k}) is the standard deviation of the surrounding elevation. The Topographic Position Index (TPI) (Figures \ref{fig:tpi10k} and \ref{fig:tpi2k}) describes whether a point is above or below the average elevation of its surroundings. The Slope (Figures \ref{fig:slope10k} and \ref{fig:slope2k}) is defined as the magnitude of the gradient vector of elevation. The West-East and South-North Derivatives (Figures \ref{fig:we_der10k}, \ref{fig:we_der2k}, \ref{fig:sn_der10k}, and \ref{fig:sn_der2k}) are the partial derivatives of elevation in the west-east and north-south directions, respectively. Further information and implementation details are provided by MeteoSwiss: {\small\url{https://github.com/MeteoSwiss/topo-descriptors}}.

\begin{figure}
  \centering
  \begin{subfigure}[t]{0.48\textwidth}
      \centering
      \includegraphics[width=\textwidth]{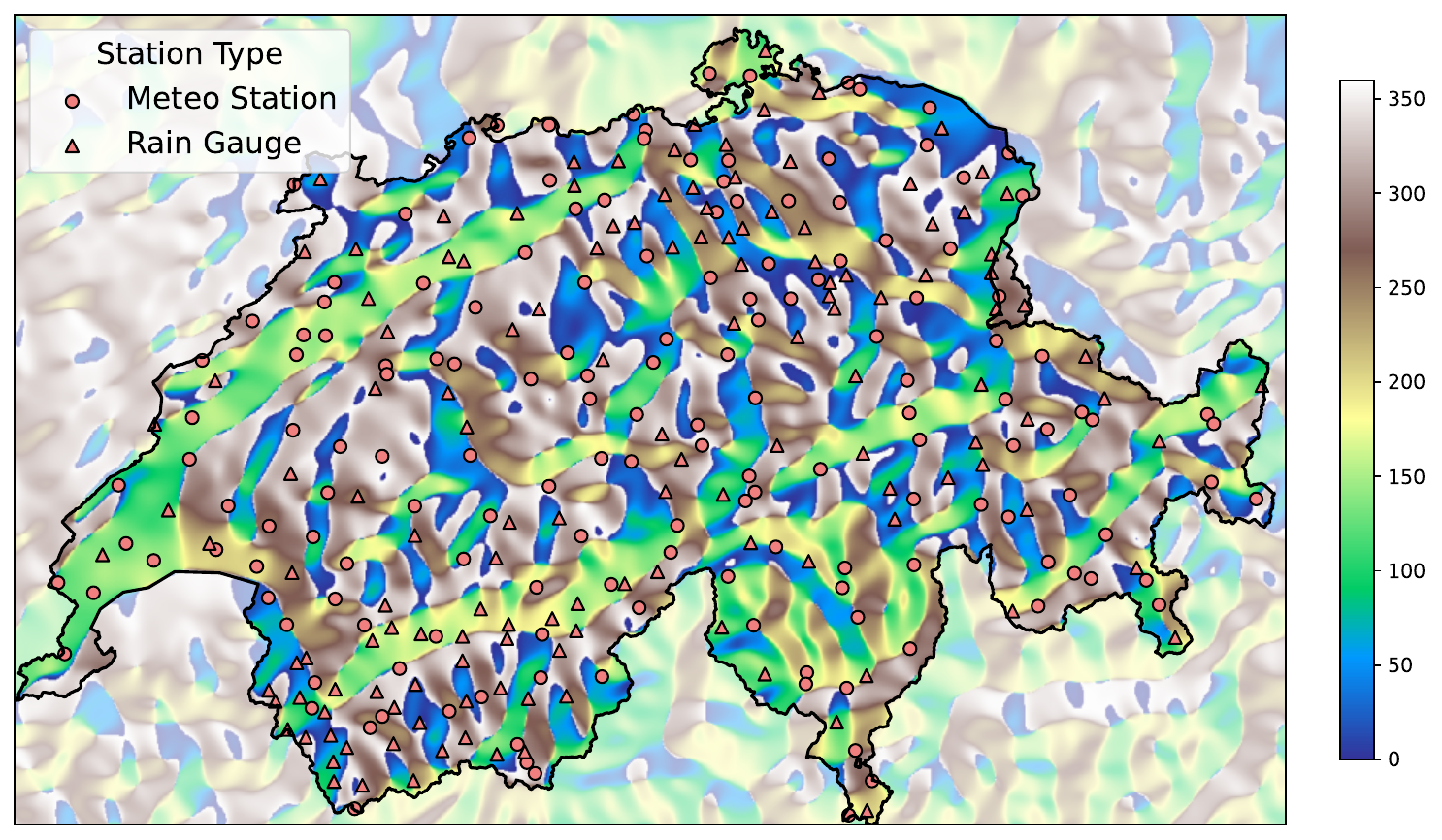}
      \caption{Aspect 10km}
      \label{fig:aspect10k}
  \end{subfigure}
  \begin{subfigure}[t]{0.48\textwidth}
      \centering
      \includegraphics[width=\textwidth]{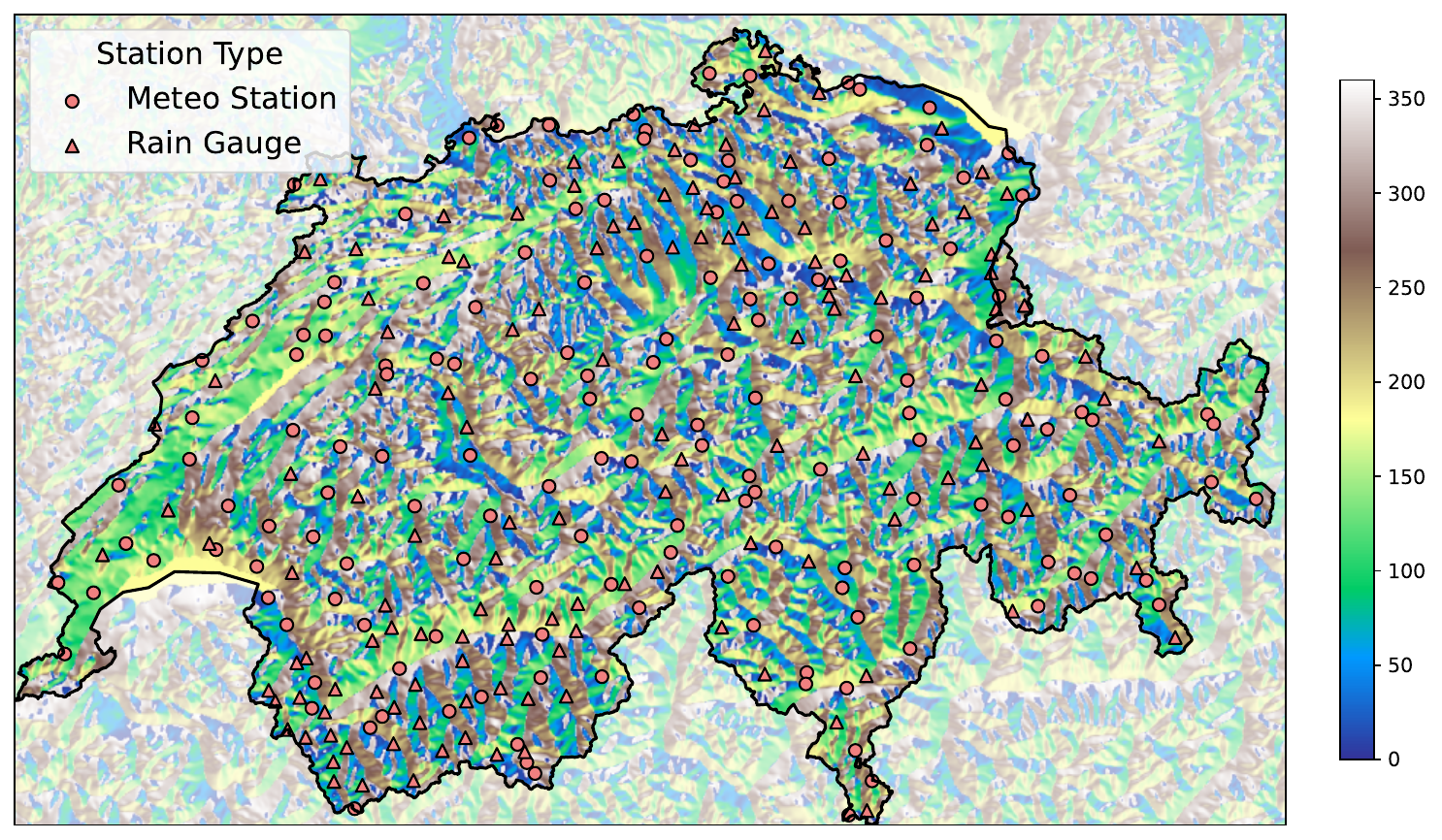}
      \caption{Aspect 2km}
      \label{fig:aspect2k}
  \end{subfigure}
  \begin{subfigure}[t]{0.48\textwidth}
      \centering
      \includegraphics[width=\textwidth]{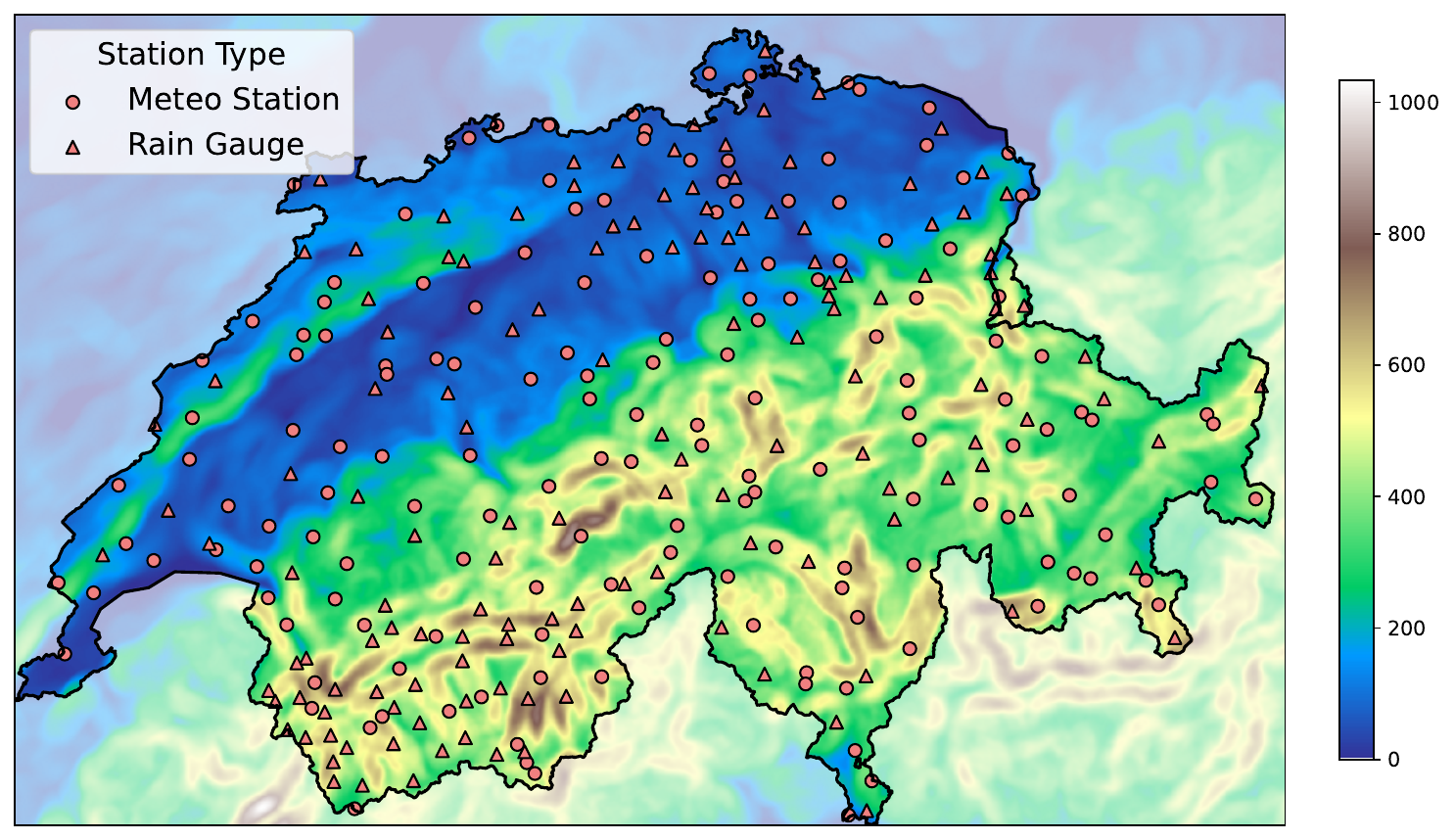}
      \caption{STD 10km}
      \label{fig:std10k}
  \end{subfigure}
  \begin{subfigure}[t]{0.48\textwidth}
      \centering
      \includegraphics[width=\textwidth]{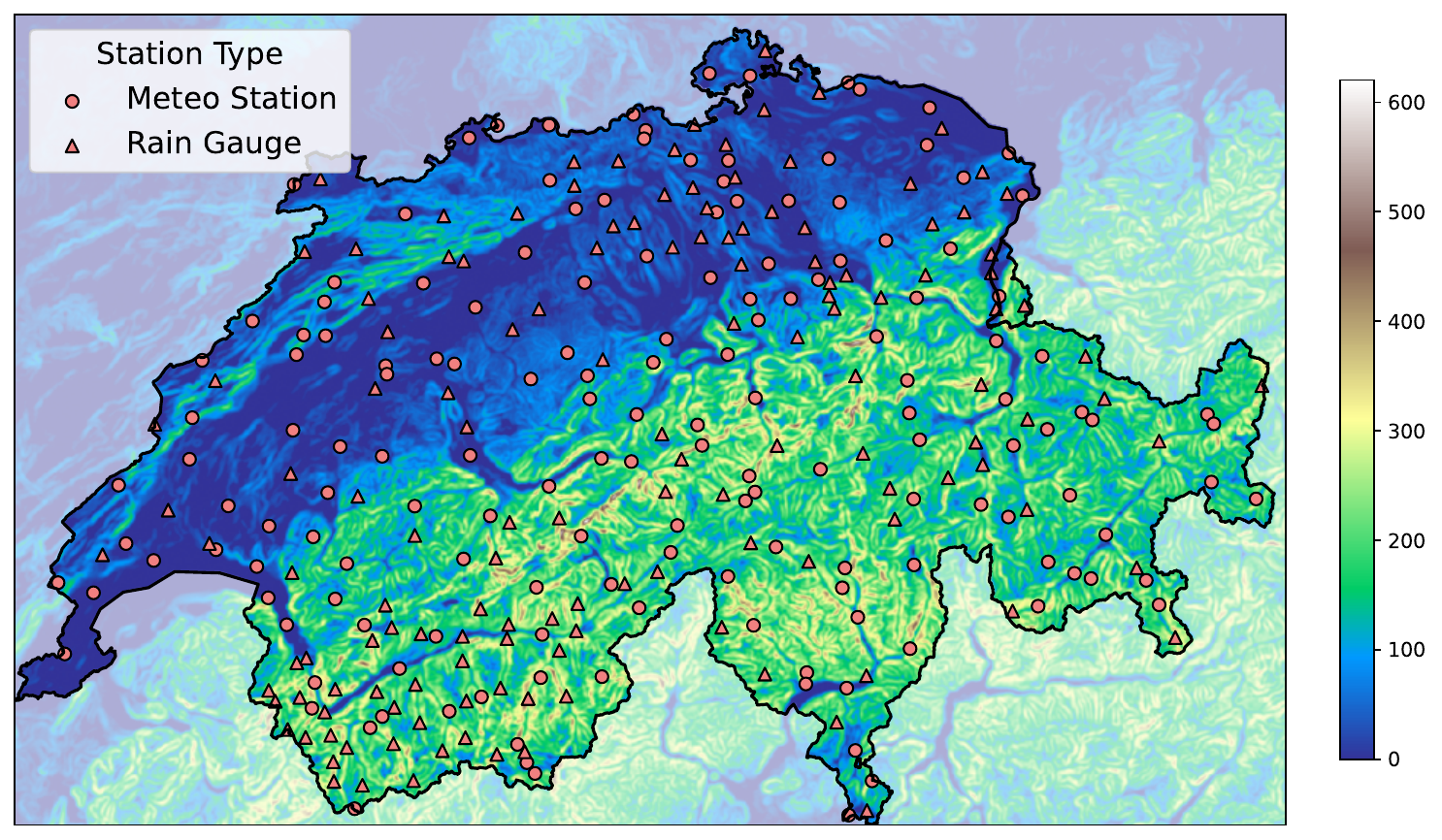}
      \caption{STD 2km}
      \label{fig:std2k}
  \end{subfigure}
  \begin{subfigure}[t]{0.48\textwidth}
      \centering
      \includegraphics[width=\textwidth]{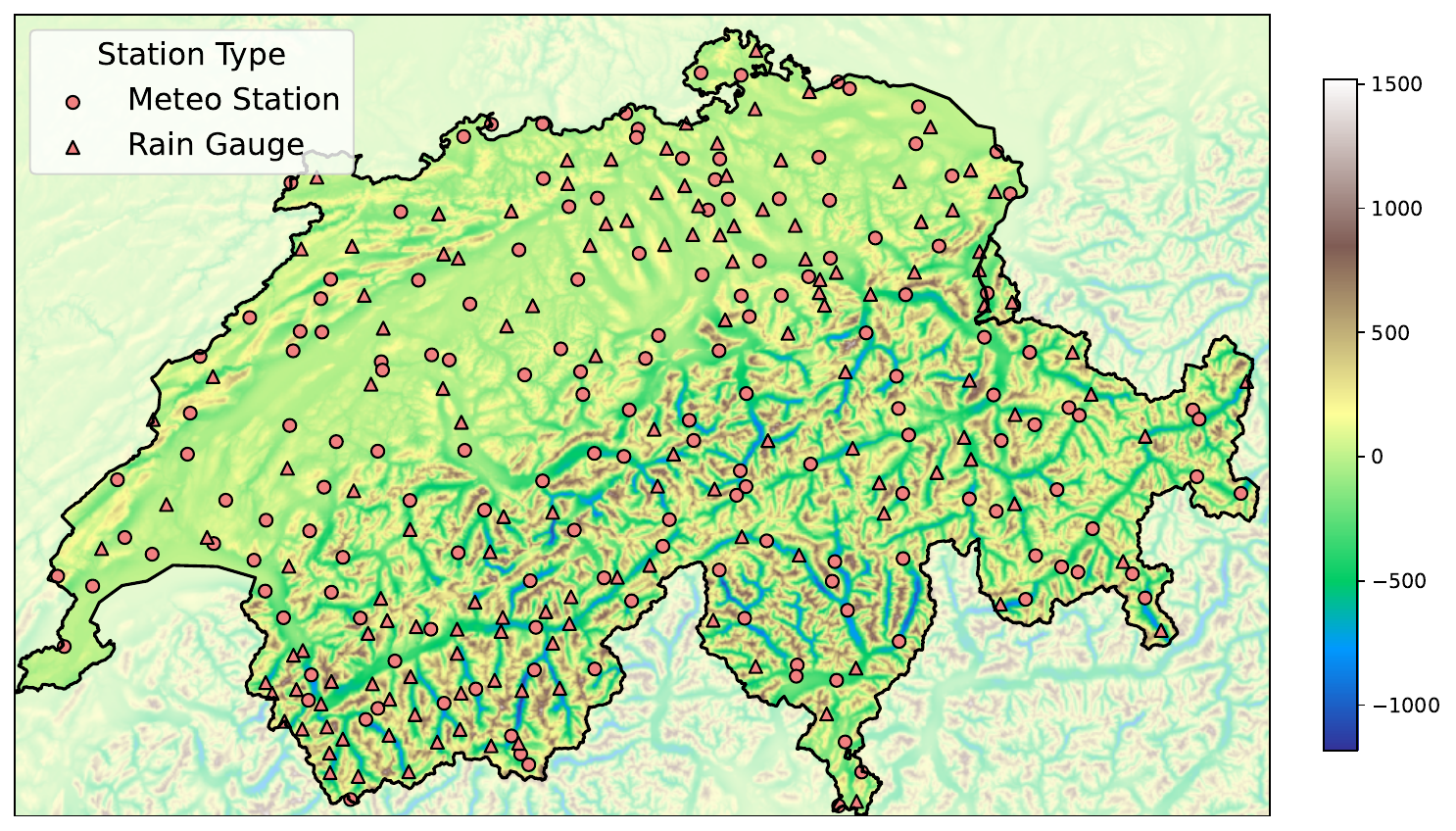}
      \caption{TPI 10km}
      \label{fig:tpi10k}
  \end{subfigure}
  \begin{subfigure}[t]{0.48\textwidth}
      \centering
      \includegraphics[width=\textwidth]{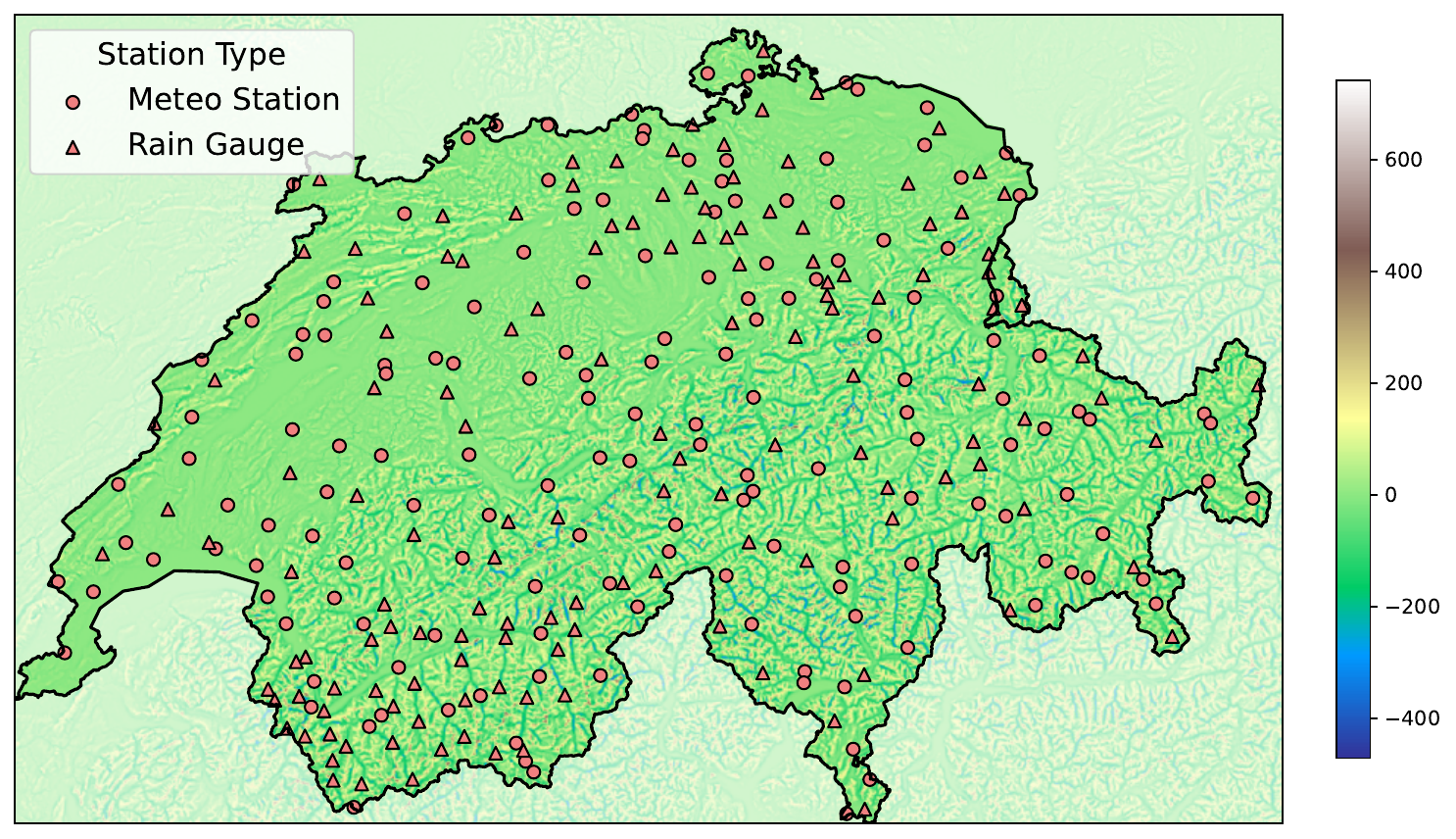}
      \caption{TPI 2km}
      \label{fig:tpi2k}
  \end{subfigure}
  \begin{subfigure}[t]{0.48\textwidth}
      \centering
      \includegraphics[width=\textwidth]{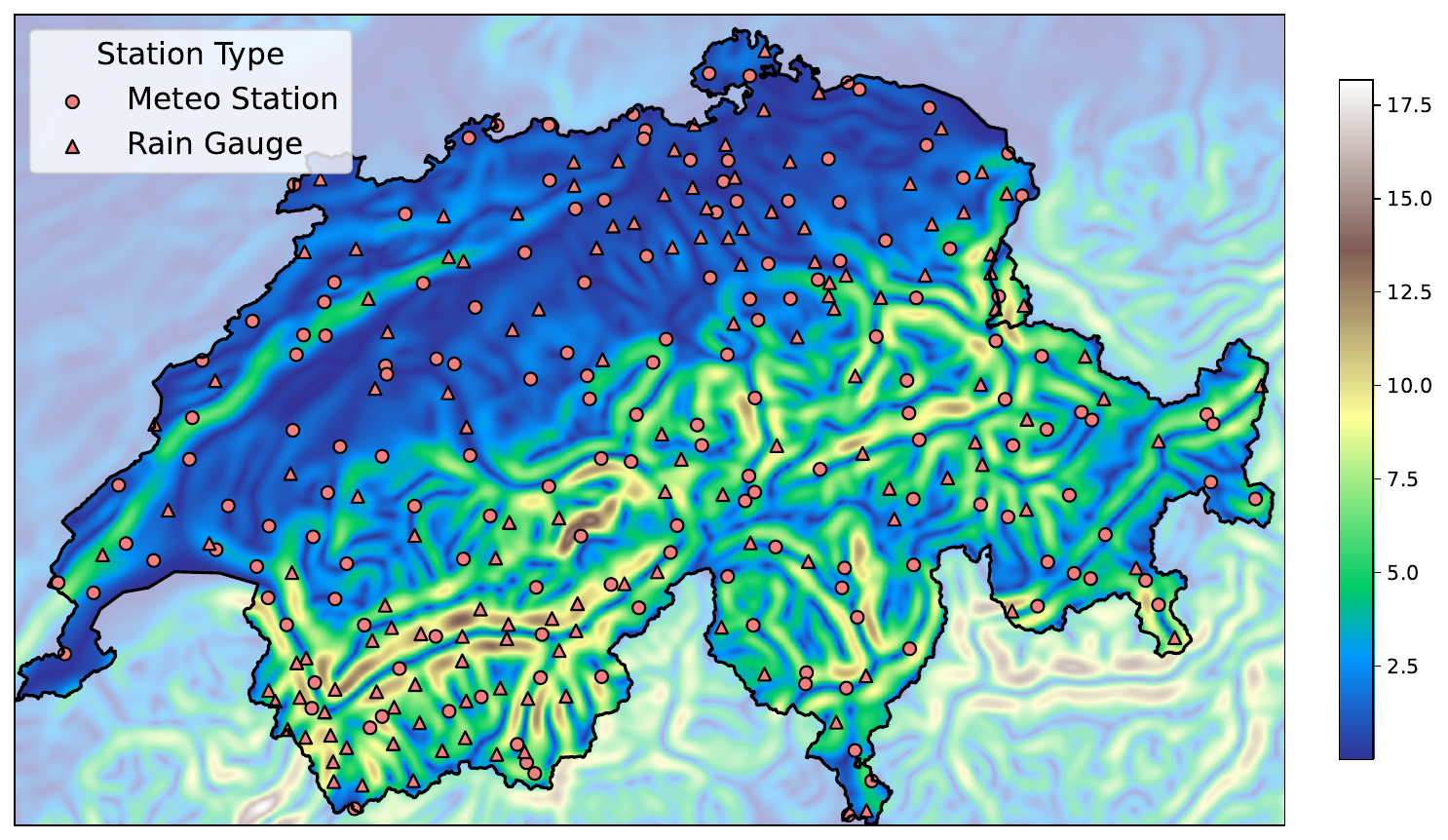}
      \caption{Slope 10km}
      \label{fig:slope10k}
  \end{subfigure}
  \begin{subfigure}[t]{0.48\textwidth}
      \centering
      \includegraphics[width=\textwidth]{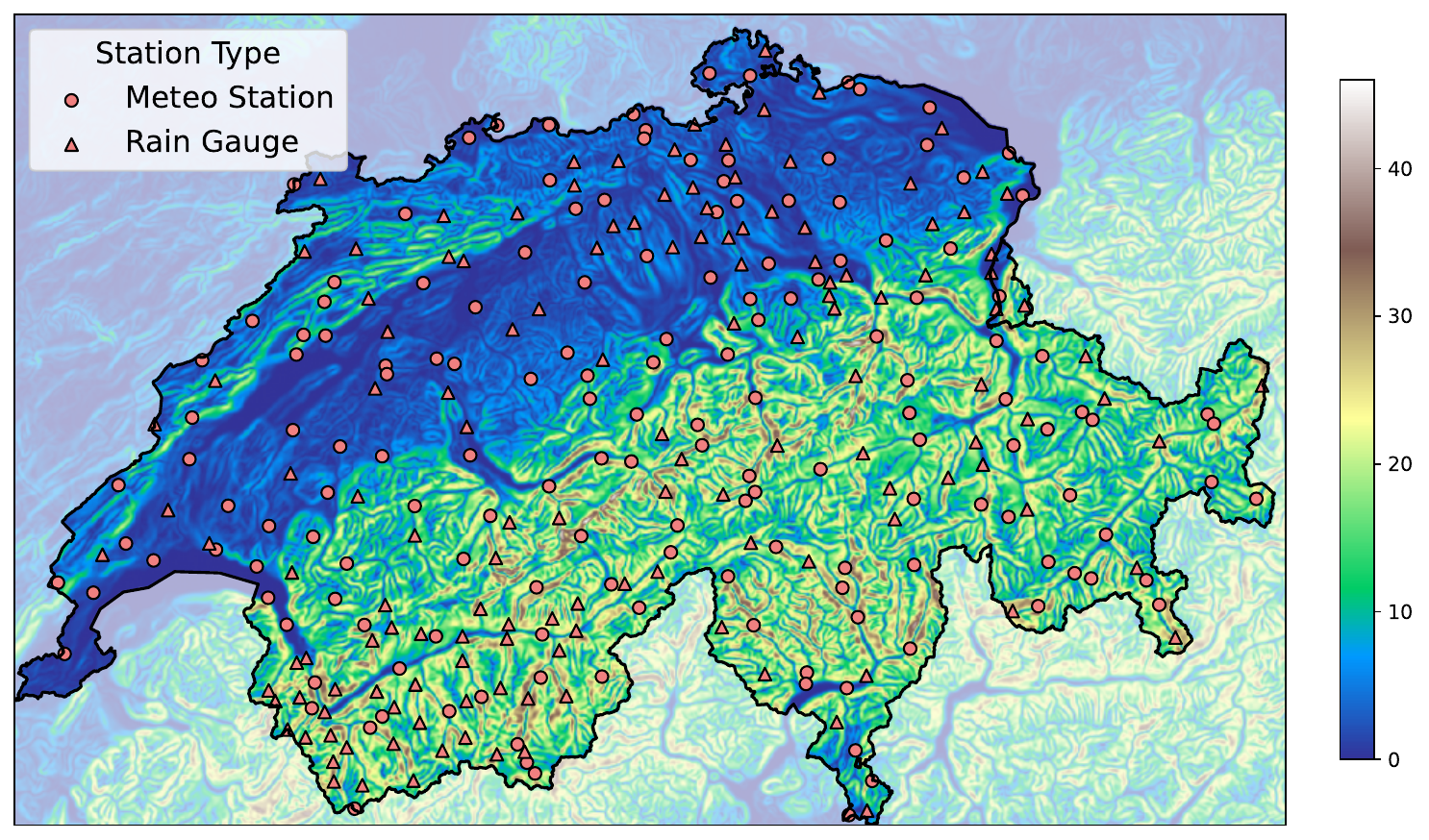}
      \caption{Slope 2km}
      \label{fig:slope2k}
  \end{subfigure}
\end{figure}

\begin{figure}[htbp]
  \ContinuedFloat
  \centering
  \begin{subfigure}[t]{0.48\textwidth}
      \centering
      \includegraphics[width=\textwidth]{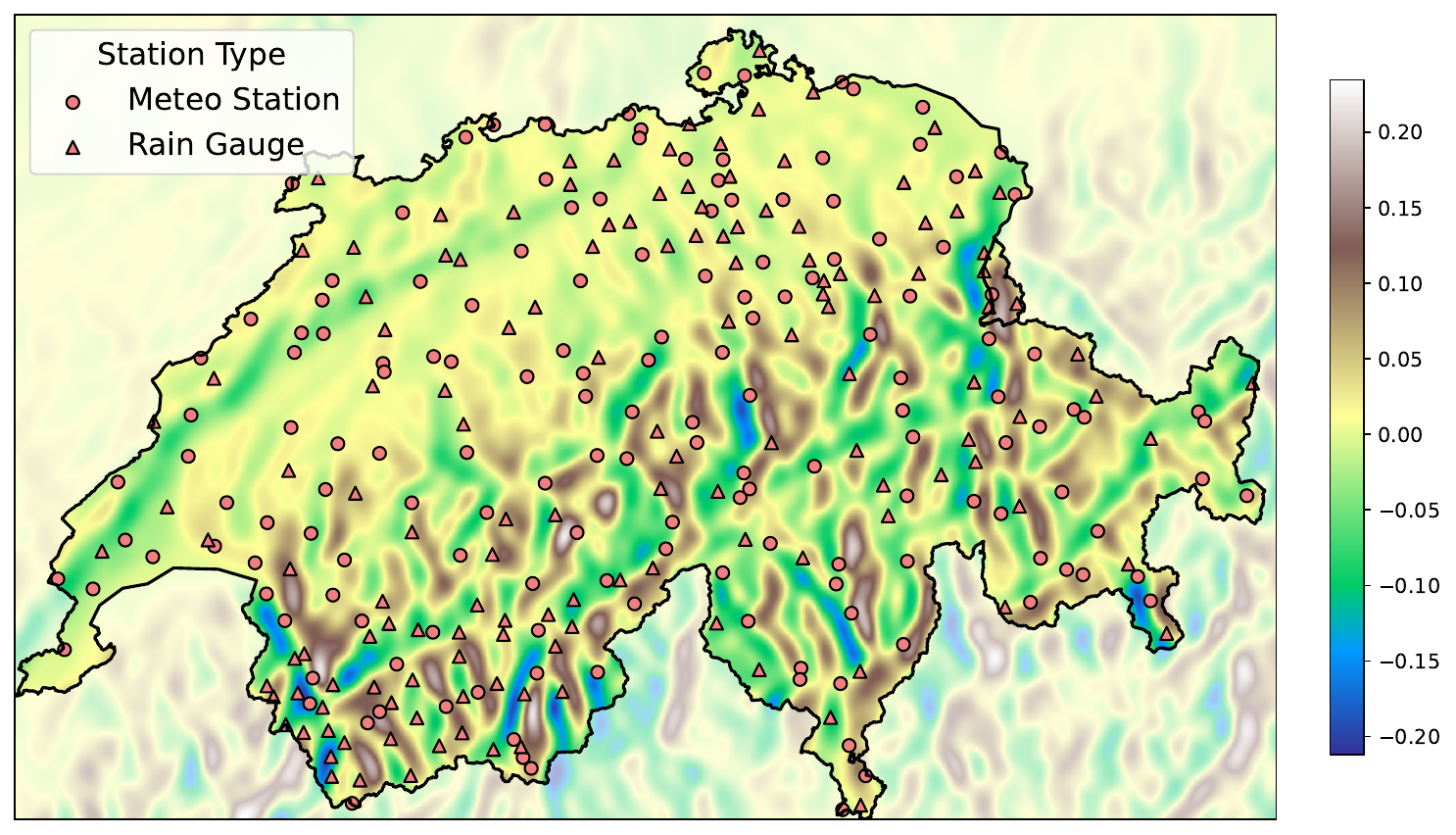}
      \caption{WE-derivative 10km}
      \label{fig:we_der10k}
  \end{subfigure}
  \begin{subfigure}[t]{0.48\textwidth}
      \centering
      \includegraphics[width=\textwidth]{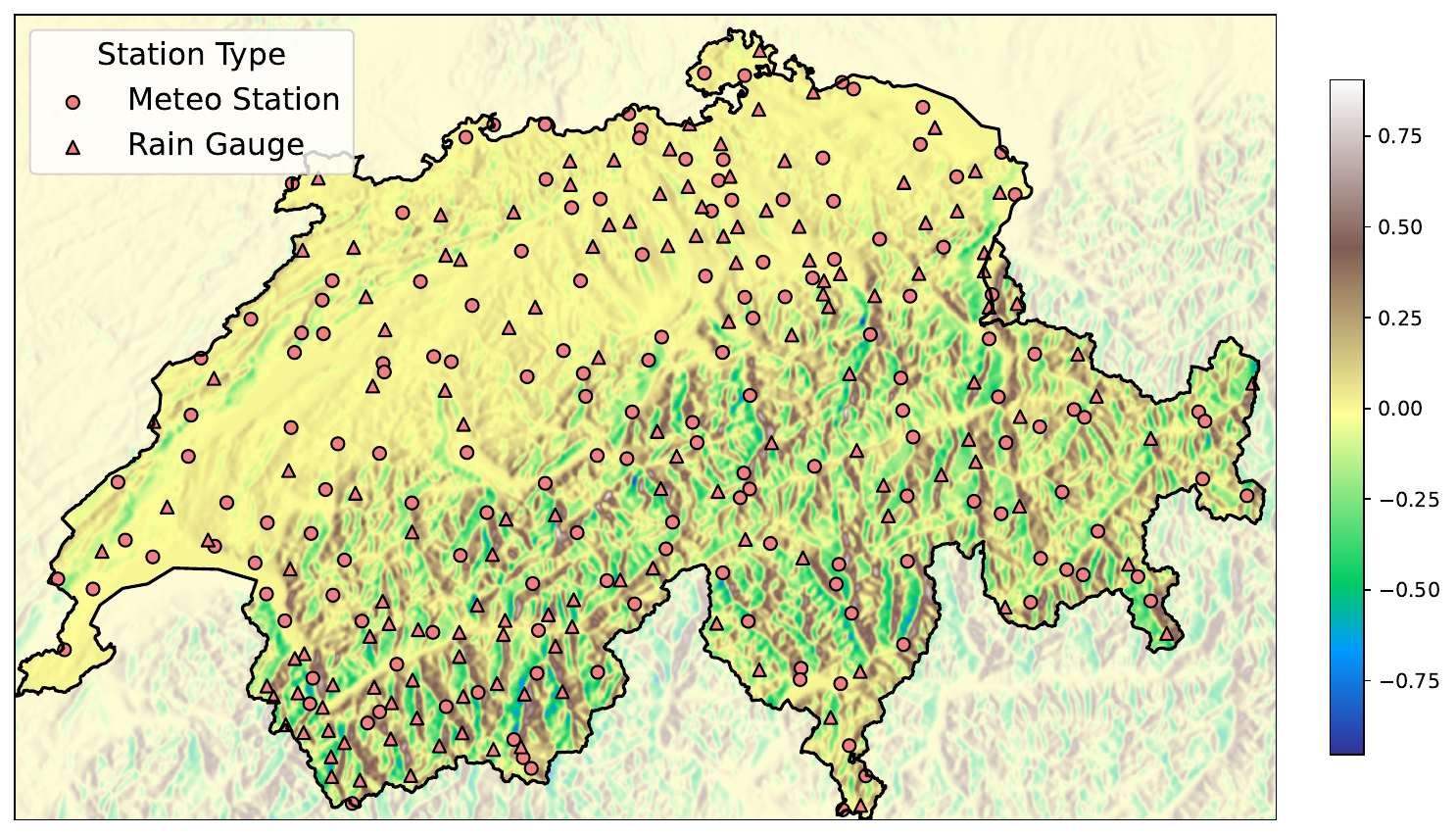}
      \caption{WE-derivative 2km}
      \label{fig:we_der2k}
  \end{subfigure}
  \begin{subfigure}[t]{0.48\textwidth}
      \centering
      \includegraphics[width=\textwidth]{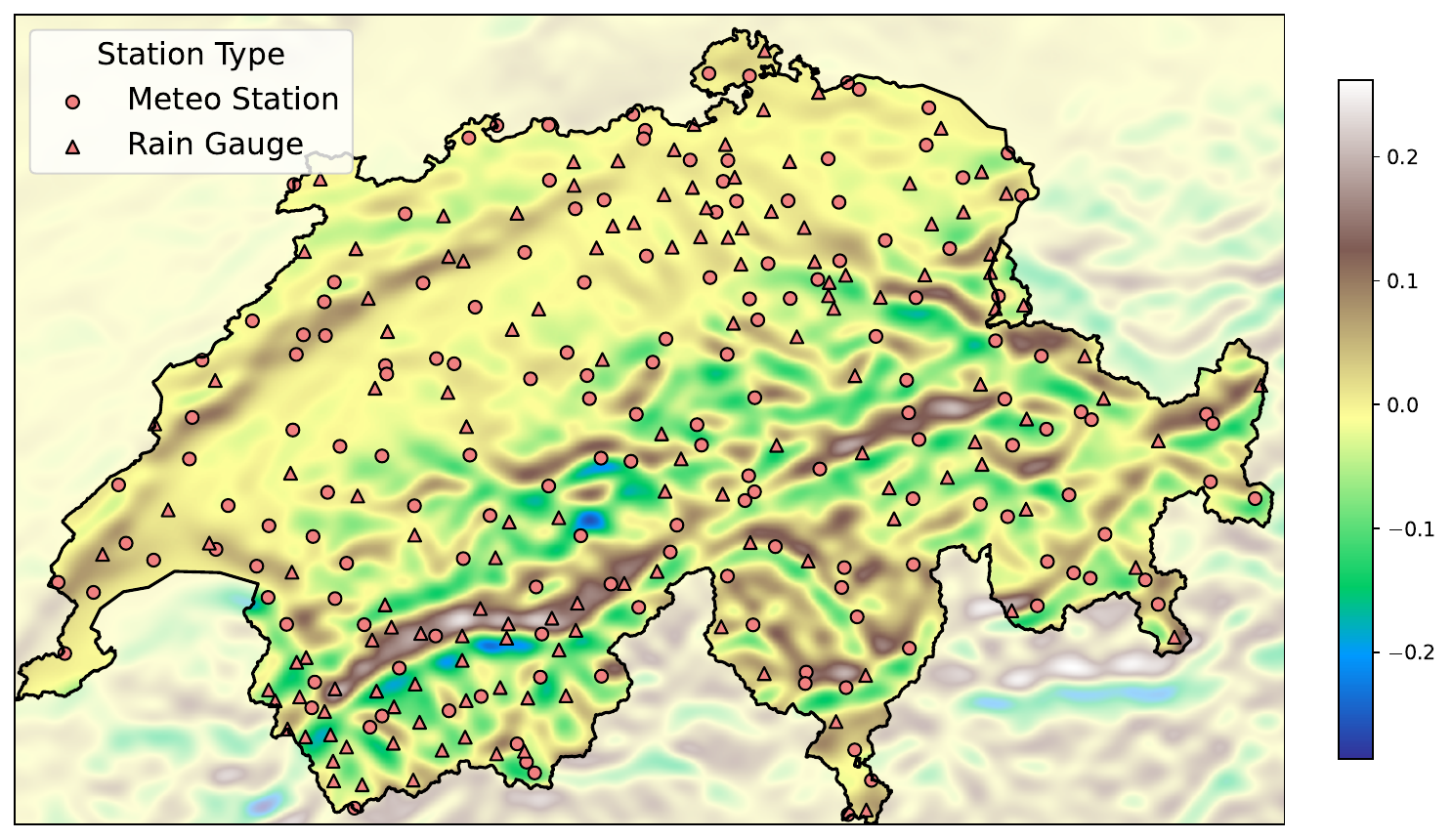}
      \caption{SN-derivative 10km}
      \label{fig:sn_der10k}
  \end{subfigure}
  \begin{subfigure}[t]{0.48\textwidth}
      \centering
      \includegraphics[width=\textwidth]{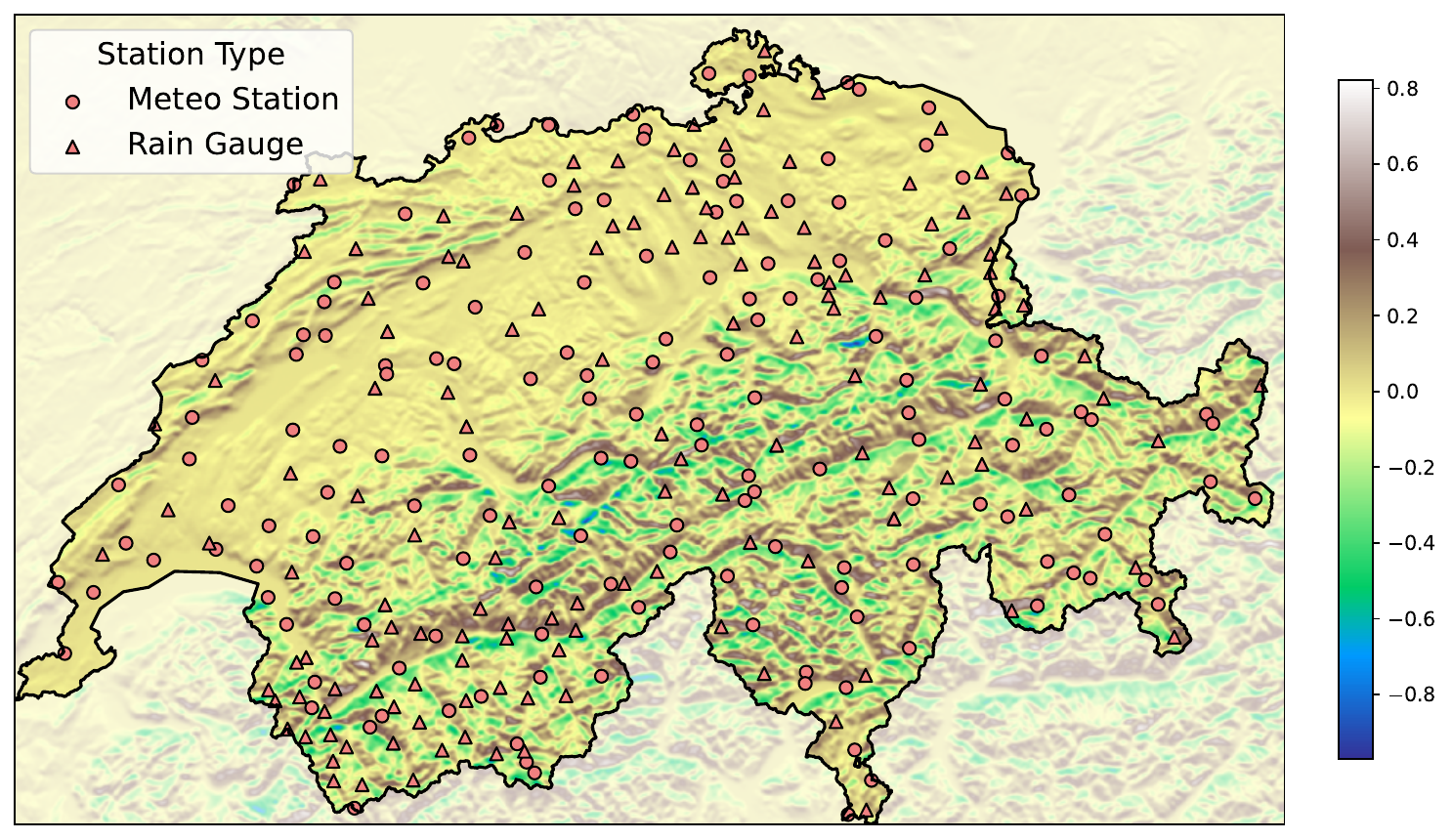}
      \caption{SN-derivative 2km}
      \label{fig:sn_der2k}
  \end{subfigure}

  \caption{Visualization of the topographic descriptors.}
  \label{fig:topo_descriptors}
\end{figure}

\section{Details about the NWP forecasting model ICON-CH1-EPS}\label{app:icon}

The numerical weather prediction (NWP) baseline from MeteoSwiss ICON-CH1-EPS consists of hourly forecasts at stations. For each forecast initialization, occurring every 3 hours, forecasts for 11 ensemble members are available. The first ensemble member is the control run, which is initialized and run with unperturbed boundary conditions (i.e., the best estimate of the state of the atmosphere). The remaining 10 ensemble members use randomly perturbed boundary conditions and stochastically perturbed physics. This also implies that there is no connection between realizations of successive initializations other than for the control member and perturbed ensemble members should be treated as statistically exchangeable.

The forecasts cover the time span from forecast initialization up to 33 hours in the future. For the aggregated quantities 
\texttt{precipitation}, \texttt{sunshine}, and \texttt{wind\_gust}, the analysis time step (i.e., forecast lead time 0) is missing. 

The ICON-CH1-EPS forecasts at stations have been extracted from the values at the nearest grid point of the regular rotated latitude-longitude grid used at MeteoSwiss for processing of high-resolution NWP data; the distance between each station and its corresponding grid point is less than 1 km. To account for the difference in model topography and altitude, temperature forecasts from ICON-CH1-EPS are altitude-corrected. A constant lapse-rate of $0.06 K/m$ is used for altitude correction.
Relative humidity is computed from temperature and dew-point temperature via the ratio of saturated vapour pressure of water derived using Equation~10 from \citep{bolton1980computation}. For the computation of relative humidity, neither temperature nor dew-point temperature is altitude corrected.

\section{Variable availability over time}\label{app:vars-availability}

\begin{figure}
    \centering
    \includegraphics[width=0.85\linewidth]{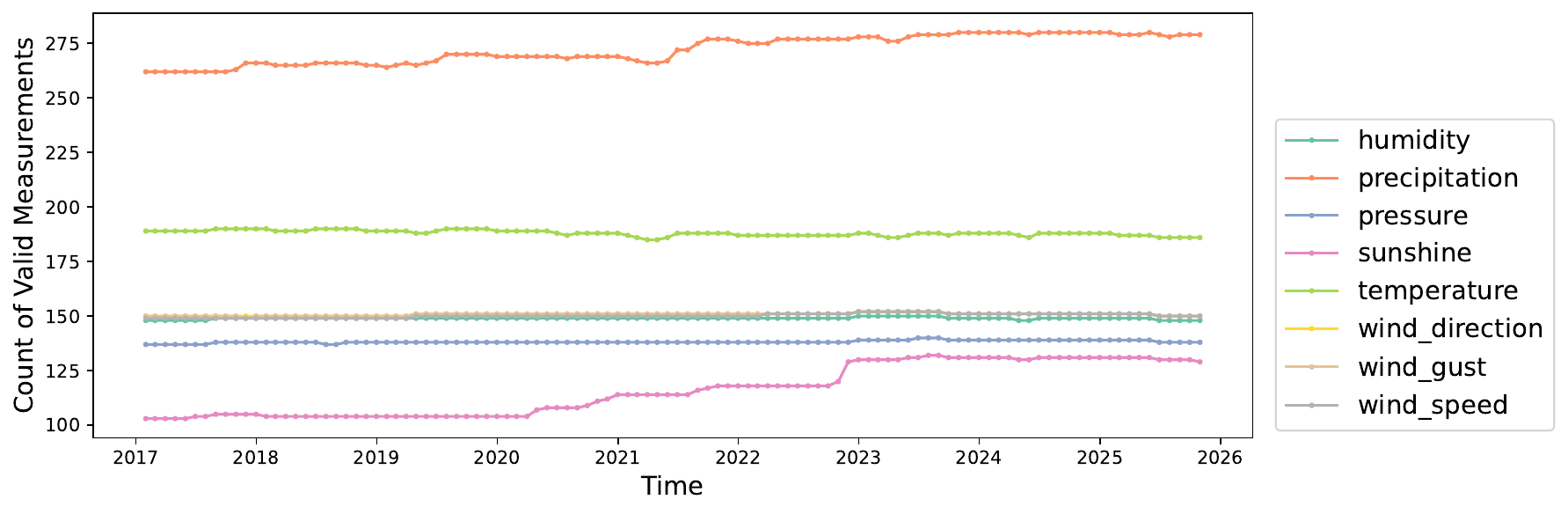}
    \caption{Number of stations with valid measurements of the meteorological variables included in PeakWeather.}
    \label{fig:missing_values}
\end{figure}

Figure \ref{fig:missing_values} displays the number of stations per month that provide valid measurements for each of the 8 meteorological variables. The plot spans the entire dataset, from January 2017 to October 2025.

\section{Additional experimental details}\label{app:exp-details}

This appendix provides additional details about the experimental setup.

\subsection{Model architectures}

The RNN models are composed of a 2-layer gated recurrent unit (GRU) with a hidden size of 128 for the temporal processing. 
The TCN models are composed of 4 temporal convolution layers using a kernel size of 3, dilation of 2, and 128 hidden units. 
The MP-TCN and MP-RNN models extend TCN and RNN, respectively, by adding 6 graph message passing layers based on diffusion convolution~\citep{li2018diffusion}, using a kernel size of 2. 
MP-TCN-gsl and MP-TCN+ variants include a graph structure learning module to respectively replace and augment the input graph used in MP-TCN with one learned from data.
MP-TCN-gsl and MP-TCN+ use 2 message passing layers over the learned graph, while MP-TCN+ also includes 4 message passing layers over the original, proximity-based graph. For these models, hidden representations from the temporal processing and the message passing blocks are added together before applying the decoder. 

The static node-level learnable embeddings $\ve^i\in \mathbb R^{D_e}$ discussed in Tables~\ref{tab:local-effects} and~\ref{tab:local-effects-gsl-extended} are learned end-to-end alongside the rest of the model parameters. As in \citep{cini2023taming}, we use them to condition the encoder and decoder, both implemented as MLPs with exponential linear unit as activation function, to account for station-specific dynamics. Their size is set to 32.
All models provide probabilistic forecasts as a Gaussian distribution (bivariate for wind forecasting and univariate for temperature forecasting, see Appendix \ref{app:temperature-extra}), for each lead time $h$, node $i$, and time step $t$. The Gaussian parameters are learned via the reparametrization trick and conditioned on the model inputs $\vx_{t-W:t}^i$, $\vu_{t-W:t}^i$, and $\vv^i$.
For PM-st and PM-day, the output distributions are $\mathcal N(\hat\vy_{t+h}^i[d], \sigma_{t-W:t}^i)$, with $\sigma_{t-W:t}^i$ being the     sampling standard deviation estimated on the time window $\vy_{t-W:t}^i[d]$.

These models are intentionally designed to be relatively straightforward deep learning models for temporal and spatiotemporal processing. The architectures and learning procedure are kept consistent across models. 
The main hyperparameters, such as the number of layers, the message passing operator, the activation, and learning rates are selected on the validation set.

Spatiotemporal models from the literature GraphWaveNet \citep{wu2019graph}, AGCRN \citep{bai2020adaptive}, and DCRNN \citep{li2018diffusion} closely follow their original implementations and hyperparameters. The learning protocol and the probabilistic readout are the same as the one used for the other models.
Chronos-2, a pre-trained 120-milion-parameter foundation model from \cite{ansari2025chronos}, is selected for its supports of multivariate and covariate-informed forecasting.
In our experiments, Chronos-2 is applied independently to each station, using the corresponding multivariate time series and associated covariates, with a substantially longer context window of 60 days that lets the model adapt to the temporal patterns of the specific timeseries. No fine-tuning was performed. Candidate window lengths ranging from 24 hours to 2 months were evaluated on the validation set.
Unlike previous models, which are sample-based, Chronos-2 produces quantiles as outputs, therefore we consider the median to assess the MAE and do not report the Energy Score.

\subsection{Graph construction and learning}
The heuristically defined graph used in the experiments is constructed based on pairwise geographical distances between stations. The Gaussian kernel with width equal to the standard deviation of the station distances is applied to compute station similarity scores.
To enforce sparsity in the graph, each node is connected to at most its top $k=8$ nearest neighbors with a similarity score exceeding 0.7.

The graph structure learning module and training used in MP-TCN-gsl and MP-TCN+ models follow \citep{cini2023sparse} to maintain computational efficiency throughout training and inference. The graph is modeled by matrix $\phi \in \mathbb R^{N\times N}$ of free model parameters that defines a distribution of top-$k$ graphs, with $k=10$. The parameters of $\phi$ are trained via score-function gradient estimation alongside the other model parameters. GraphWaveNet and AGCRN implement different graph structure learning strategies.

\subsection{Model training and evaluation}
Model predictions are made for a 24-hour horizon from an input window of 24 time steps (1 day). All weather variables, encodings of the hour of the day and the hour of the year, the binary mask of available observations, and all provided topographic descriptors are considered as input features to the model. 
Models are trained on data up to September 2024 by optimizing the Energy Score (ES) \citep{gneiting2007strictly} at every station and every lead time (up to 24 hours ahead) and tested on the last year, appropriately masking missing observations. 
Validated is performed on approximately one fifth of data covering October 2023 to September 2024. 
The Energy Score is here estimated via Monte Carlo 
\begin{equation*}
    \text{ES}\left(\left\{\hat \vy^{i,(m)}_{t+h}\right\},\vy_{t+h}^i\right) = 
    \frac{\sum_{m=1}^M \big\lVert \hat \vy^{i,(m)}_{t+h} - \vy_{t+h}^i\big\rVert_2}{M}  
    -
    \frac{ \sum_{m>l=1}^M \big\lVert \hat \vy^{i,(m)}_{t+h} - \hat \vy_{t+h}^{i,(l)}\big\rVert_2 }{M (M-1)},
\end{equation*}
with $\vy_{t+h}^i$ the target vector (such as wind velocity) and $\{\hat \vy^{i,(m)}_{t+h}\}_{m=1}^M$ samples from the model's predicted distribution. During training, $M=16$ samples are considered, while $M=100$ for testing.
When training MP-TCN+ and MP-TCN-gsl, a new graph is sampled for each batch, while, at inference time, the model relies on the mode of the distribution. 
Training is performed for 200 epochs using Adam optimizer \cite{adam2014method} with an initial learning rate of 0.001 decreased at epochs 40, 60 and 120 by a factor of 0.3. Each epoch consists of 300 batches of size 32. Early stopping is triggered after 40 epochs without improvement on the validation forecasting MAE.
Models are evaluated by averaging forecasting performance over the 24h horizon and evaluating at specific lead times. To assess the quality of wind direction forecasts, samples where the corresponding wind speed are below 1m/s are discarded.

As ICON issues predictions every 3 hours, we consider two test sets of data from October 2024. One consists only of the windows for which NWP forecasts are available and the other contains all possible windows. Unless specified otherwise, the former is considered to compare with ICON. Table~\ref{tab:wind} shows that the metrics are consistent across the two test sets.

\subsection{Hardware and Software}
Experiments are run on a machine with Intel(R) Xeon(R) CPU, 192 GB RAM, and NVIDIA L4 GPUs. The system runs Debian GNU/Linux 11.
The code to run the experiments is developed in Python~\citep{rossum2009python}, by relying mainly on the following open-source libraries: PyTorch~\citep{paske2019pytorch}, PyTorch Lightning~\citep{Falcon_PyTorch_Lightning_2019}, PyTorch Geometric (PyG) \citep{fey2019fast}, Torch Spatiotemporal~\citep{Cini_Torch_Spatiotemporal_2022}, and the developed PeakWeather library \citep{PeakWeatherLibraryBaseCode} to interface with the PeakWeather dataset \citep{PeakWeatherDataset}. The source code with all the configuration files necessary to reproduce the results is open-sourced on GitHub.%
    \footnote{\url{https://github.com/Graph-Machine-Learning-Group/peakweather-forecasting}}
Model training and evaluation do not require particular hardware. Given the above machine and experimental settings, single model training and evaluation should complete within 3 hours.

\section{Extended results for wind forecasting}\label{app:wind-extra}

To complement the empirical results of Table~\ref{tab:local-effects} and Figure~\ref{fig:wind-1d}, we report here additional results from the recurrent architectures and additional models from the literature.

Table~\ref{tab:local-effects-gsl-extended} studies the conditioning of model predictions on station-specific components and relational information. Results for the RNN, MP-RNN, and GraphWaveNet confirm those in Table~\ref{tab:local-effects}. In Table~\ref{tab:local-effects-gsl-extended}, GraphWaveNet-nogsl refers to the same GraphWaveNet architecture excluding the module that learns a graph in addition to the input one. 

Table~\ref{tab:wind} instead compares the MAE on the velocity, speed, and direction, and Energy Score aggregated over the entire forecasting horizon. RNN-noemb refers to the RNN without learnable embeddings. Results confirm the superior performance of graph-based models in this setting and show consistent results across the two test sets.

\begin{table}
\caption{Impact on prediction performance of accounting for station-specific features and graph structures. The last columns indicate the size $D_e$ of the learnable node embeddings used (Emb.), and whether the corresponding model relies on station-specific topographic descriptors $\vv_i$ (Topo.), an input graph (In.G.), and a graph structure learned from the data (GSL). Results are aggregated over five runs with different random seeds and are reported as mean $\pm$ standard deviation. Symbol ``-'' indicates non-applicable entries. For all metrics, the lower the better. 
}
\label{tab:local-effects-gsl-extended}
\centering
\begin{tabular}{lllllcccc}
\toprule
Model & MAE Vel. & MAE Speed & MAE Dir. & En.Score & Emb. & Topo. & In.G. & GSL \\
\midrule
RNN          & 1.122\textsubscript{$\pm$0.004} & 1.070\textsubscript{$\pm$0.004} & 44.656\textsubscript{$\pm$0.382} & 1.274\textsubscript{$\pm$0.006} & 0  & \abse & -  & - \\
RNN          & 1.085\textsubscript{$\pm$0.003} & 1.037\textsubscript{$\pm$0.001} & 41.848\textsubscript{$\pm$0.129} & 1.226\textsubscript{$\pm$0.003} & 0  & \pres & -  & -  \\
RNN          & 1.092\textsubscript{$\pm$0.003} & 1.042\textsubscript{$\pm$0.003} & 42.352\textsubscript{$\pm$0.109} & 1.233\textsubscript{$\pm$0.003} & 32 & \abse & -  & -  \\
RNN          & 1.087\textsubscript{$\pm$0.002} & 1.040\textsubscript{$\pm$0.002} & 42.050\textsubscript{$\pm$0.175} & 1.226\textsubscript{$\pm$0.002} & 32 & \pres & -  & -  \\
\midrule
MP-RNN       & 1.060\textsubscript{$\pm$0.005} & 1.022\textsubscript{$\pm$0.004} & 39.990\textsubscript{$\pm$0.217} & 1.198\textsubscript{$\pm$0.006} & 0  & \abse & \pres  & \abse  \\
MP-RNN       & 1.056\textsubscript{$\pm$0.002} & 1.019\textsubscript{$\pm$0.002} & 39.830\textsubscript{$\pm$0.194} & 1.198\textsubscript{$\pm$0.004} & 0  & \pres & \pres  & \abse  \\
MP-RNN       & 1.058\textsubscript{$\pm$0.009} & 1.022\textsubscript{$\pm$0.007} & 40.146\textsubscript{$\pm$0.762} & 1.197\textsubscript{$\pm$0.008} & 32 & \abse & \pres  & \abse  \\
MP-RNN       & 1.056\textsubscript{$\pm$0.008} & 1.021\textsubscript{$\pm$0.007} & 40.027\textsubscript{$\pm$0.641} & 1.196\textsubscript{$\pm$0.008} & 32 & \pres & \pres  & \abse  \\
\midrule
TCN          & 1.112\textsubscript{$\pm$0.002} & 1.062\textsubscript{$\pm$0.001} & 44.370\textsubscript{$\pm$0.202} & 1.262\textsubscript{$\pm$0.002} & 0  & \abse & -  & -  \\
TCN          & 1.077\textsubscript{$\pm$0.003} & 1.030\textsubscript{$\pm$0.002} & 41.644\textsubscript{$\pm$0.182} & 1.215\textsubscript{$\pm$0.004} & 0  & \pres & -  & -  \\
TCN          & 1.076\textsubscript{$\pm$0.001} & 1.029\textsubscript{$\pm$0.001} & 41.607\textsubscript{$\pm$0.091} & 1.213\textsubscript{$\pm$0.001} & 32 & \abse & -  & -  \\
TCN          & 1.076\textsubscript{$\pm$0.002} & 1.028\textsubscript{$\pm$0.001} & 41.577\textsubscript{$\pm$0.223} & 1.212\textsubscript{$\pm$0.002} & 32 & \pres & -  & -  \\
\midrule
MP-TCN       & 1.024\textsubscript{$\pm$0.004} & 0.989\textsubscript{$\pm$0.003} & 38.468\textsubscript{$\pm$0.255} & 1.154\textsubscript{$\pm$0.005} & 0  & \abse & \pres & \abse  \\
MP-TCN       & 1.014\textsubscript{$\pm$0.003} & 0.981\textsubscript{$\pm$0.003} & 37.857\textsubscript{$\pm$0.091} & 1.141\textsubscript{$\pm$0.003} & 0  & \pres & \pres & \abse  \\
MP-TCN       & 1.011\textsubscript{$\pm$0.001} & 0.979\textsubscript{$\pm$0.001} & 37.795\textsubscript{$\pm$0.136} & 1.137\textsubscript{$\pm$0.002} & 32 & \abse & \pres & \abse  \\
MP-TCN       & 1.010\textsubscript{$\pm$0.002} & 0.978\textsubscript{$\pm$0.002} & 37.676\textsubscript{$\pm$0.117} & 1.135\textsubscript{$\pm$0.002} & 32 & \pres & \pres & \abse  \\
MP-TCN-gsl   & 1.001\textsubscript{$\pm$0.003} & 0.973\textsubscript{$\pm$0.003} & 37.214\textsubscript{$\pm$0.173} & 1.125\textsubscript{$\pm$0.004} & 32 & \pres & \abse & \pres \\
MP-TCN+      & 0.999\textsubscript{$\pm$0.001} & 0.971\textsubscript{$\pm$0.002} & 37.19\textsubscript{$\pm$0.12}   & 1.125\textsubscript{$\pm$0.002} & 32 & \pres & \pres & \pres \\
\midrule
GraphWaveNet-nogsl  & 1.011\textsubscript{$\pm$0.003} & 0.977\textsubscript{$\pm$0.002} & 37.789\textsubscript{$\pm$0.155} & 1.134\textsubscript{$\pm$0.003} & -  & \pres & \pres & \abse \\
GraphWaveNet        & 0.994\textsubscript{$\pm$0.001} & 0.964\textsubscript{$\pm$0.001} & 36.85\textsubscript{$\pm$0.10}   & 1.116\textsubscript{$\pm$0.001} & -  & \pres & \pres & \pres \\
\bottomrule
\end{tabular}
\end{table}

\begin{table}
\caption{Performance comparison of deep learning models with ICON and persistence model baselines for wind forecasting. Results report average $\pm$ 1 standard deviation computed over five runs with different random seeds, except for ICON and Chronos-2: ICON provides a single set of ensemble forecasts, while Chronos-2 predicted quantiles are not affected by initialization. As the persistence models require no training, their variability arises solely from Monte Carlo sampling during metric evaluation. The top part of the table considers the NWP test set with forecasts issued every 3h to compare with ICON, while the bottom part the full held-out test data.}
\label{tab:wind}
\centering
\begin{tabular}{lllllcc}
\toprule
Model & MAE Velocity & MAE Speed & MAE Direction & Energy Score & Input Graph & Learn Graph\\
\midrule
\multicolumn{7}{c}{\emph{NWP Test Set}}\\
\midrule
RNN-noemb & 1.090\textsubscript{$\pm$0.001} & 1.040\textsubscript{$\pm$0.001} & 42.03\textsubscript{$\pm$0.11} & 1.229\textsubscript{$\pm$0.001} & - & - \\
RNN & 1.087\textsubscript{$\pm$0.002} & 1.040\textsubscript{$\pm$0.002} & 42.05\textsubscript{$\pm$0.18} & 1.226\textsubscript{$\pm$0.002} & - & - \\
TCN & 1.075\textsubscript{$\pm$0.001} & 1.027\textsubscript{$\pm$0.001} & 41.46\textsubscript{$\pm$0.05} & 1.211\textsubscript{$\pm$0.000} & - & - \\
MP-RNN & 1.056\textsubscript{$\pm$0.008} & 1.021\textsubscript{$\pm$0.007} & 40.03\textsubscript{$\pm$0.64} & 1.196\textsubscript{$\pm$0.008} & \pres & \abse \\
MP-TCN & 1.012\textsubscript{$\pm$0.004} & 0.980\textsubscript{$\pm$0.003} & 37.78\textsubscript{$\pm$0.20} & 1.137\textsubscript{$\pm$0.004} & \pres & \abse \\
MP-TCN+ & 0.999\textsubscript{$\pm$0.001} & 0.971\textsubscript{$\pm$0.002} & 37.19\textsubscript{$\pm$0.12} & 1.125\textsubscript{$\pm$0.002} & \pres & \pres \\
\midrule
PM-day & 1.589\textsubscript{$\pm$0.000} & 1.524\textsubscript{$\pm$0.000} & 57.97\textsubscript{$\pm$0.01} & 1.921\textsubscript{$\pm$0.000} & - & - \\
PM-st & 1.519\textsubscript{$\pm$0.000} & 1.472\textsubscript{$\pm$0.000} & 60.16\textsubscript{$\pm$0.00} & 1.819\textsubscript{$\pm$0.000} & - & - \\
ICON & 1.132 & 1.113 & 39.073 & 1.407  & - & - \\
\midrule
DCRNN & 1.049\textsubscript{$\pm$0.003} & 1.013\textsubscript{$\pm$0.002} & 39.75\textsubscript{$\pm$0.17} & 1.182\textsubscript{$\pm$0.003} & \pres & - \\
AGCRN & 1.038\textsubscript{$\pm$0.003} & 1.005\textsubscript{$\pm$0.003} & 39.10\textsubscript{$\pm$0.20} & 1.172\textsubscript{$\pm$0.005} & - & \pres \\
GraphWaveNet & 0.994\textsubscript{$\pm$0.001} & 0.964\textsubscript{$\pm$0.001} & 36.85\textsubscript{$\pm$0.10} & 1.116\textsubscript{$\pm$0.001} & \pres & \pres \\
Chronos-2 & 1.108 & 1.202 & 42.96 & - & - & - \\
\toprule
\multicolumn{7}{c}{\emph{Full Test Set}}\\
\toprule
RNN-noemb    & 1.088\textsubscript{$\pm$0.001} & 1.038\textsubscript{$\pm$0.001} & 41.95\textsubscript{$\pm$0.09} & 1.227\textsubscript{$\pm$0.001} & - & - \\
RNN          & 1.087\textsubscript{$\pm$0.001} & 1.038\textsubscript{$\pm$0.002} & 42.11\textsubscript{$\pm$0.23} & 1.225\textsubscript{$\pm$0.002} & - & - \\
TCN          & 1.074\textsubscript{$\pm$0.001} & 1.026\textsubscript{$\pm$0.001} & 41.42\textsubscript{$\pm$0.05} & 1.209\textsubscript{$\pm$0.000} & - & - \\
MP-RNN       & 1.056\textsubscript{$\pm$0.008} & 1.019\textsubscript{$\pm$0.007} & 40.03\textsubscript{$\pm$0.63} & 1.195\textsubscript{$\pm$0.008} & \pres & \abse \\
MP-TCN       & 1.012\textsubscript{$\pm$0.004} & 0.978\textsubscript{$\pm$0.002} & 37.78\textsubscript{$\pm$0.20} & 1.136\textsubscript{$\pm$0.004} & \pres & \abse \\
MP-TCN+      & 0.999\textsubscript{$\pm$0.001} & 0.969\textsubscript{$\pm$0.002} & 37.19\textsubscript{$\pm$0.12} & 1.123\textsubscript{$\pm$0.002} & \pres & \pres \\
\midrule
PM-day     & 1.586\textsubscript{$\pm$0.000} & 1.522\textsubscript{$\pm$0.000} & 57.87\textsubscript{$\pm$0.00} & 1.917\textsubscript{$\pm$0.000} & - & - \\
PM-st      & 1.519\textsubscript{$\pm$0.000} & 1.471\textsubscript{$\pm$0.000} & 60.20\textsubscript{$\pm$0.00} & 1.818\textsubscript{$\pm$0.000} & - & - \\
\bottomrule
\end{tabular}
\end{table}

\section{Temperature forecasting: an inductive setup}\label{app:temperature-extra}

We present an additional experiment on an inductive learning task for temperature prediction at previously unseen locations. When loading the dataset, both meteorological stations and rain gauges were considered. 
We consider all the stations equipped with a temperature sensor and do not use any covariates. Of those stations, 20 meteorological stations and 20 rain gauges are removed from the training set and used only for testing. 
This setting limits the breadth of models that can be tested, as graph structure learning and learned station embeddings cannot straightforwardly transfer to new nodes. 
Static attributes related to the station location, such as elevation and topographic features can instead be used. 
Besides these variations, the same hyperparameters used for wind forecasting are also considered here.

\begin{figure}
\centering
\includegraphics[width=\textwidth]{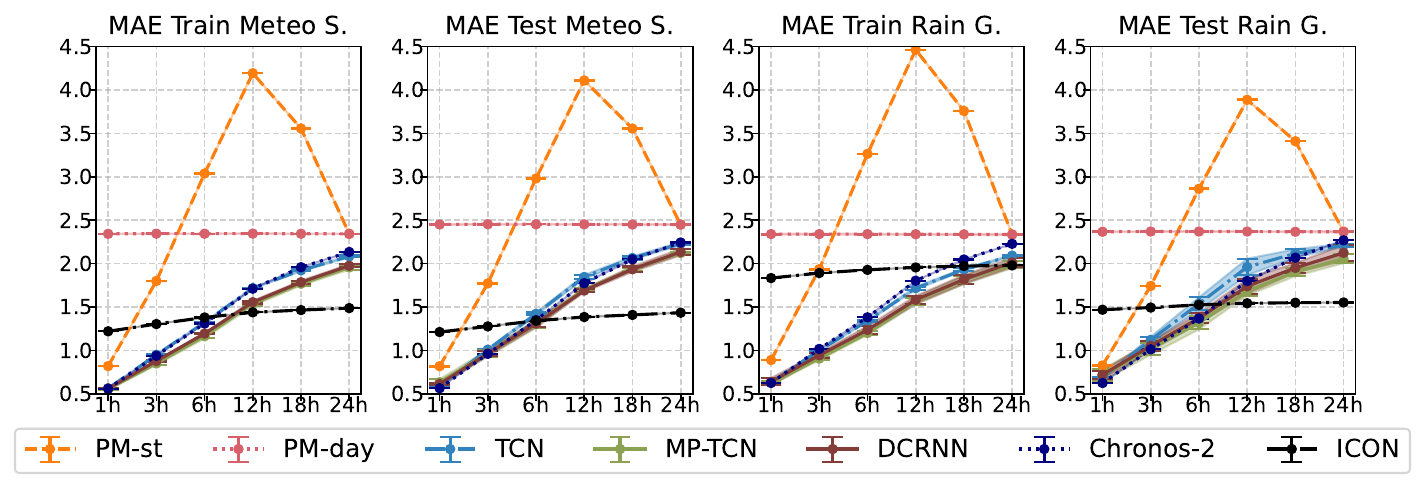}
\caption{Performance comparison of deep learning models with ICON, persistence model baselines and Chronos-2 for inductive temperature forecasting. Results are averaged over five runs with different random seeds, except for ICON and Chronos-2.  
Shaded areas indicate $\pm$3 standard deviations. From left to right, the four plots represent the results obtained on seen meteorological stations, unseen meteorological stations, seen rain gauges, and unseen rain gauges.}
\label{fig:temp}
\end{figure}

Table~\ref{tab:temp} summarizes the results in terms of MAE and ES for the various trainable models, persistence models, the foundation model Chronos-2, and ICON. Results are reported separately for each type of station (all, meteorological only, and rain gauge only) and further distinguished between results obtained on stations included in the training set (seen stations) and results obtained on the held-out set (unseen stations).
First we comment that the performance is sensitive to the specific choice of held-out stations. We can see it from the difference in metrics between seen and unseen stations of the models of the first block (PM-st, PM-day, ICON, and Chronos-2), which are not trained on the seen/training stations; such sensitivity was observed---although not shown here---across different sets of held-out stations. For trained models (the second block with TCN, MP-TCN, and DCRNN) the predictions degrade slightly, overall by 10\%. However, we note that for the spatiotemporal models, the performance is still significantly better than that of ICON and Chronos-2. Figure \ref{fig:temp} illustrates the performance across lead times.
Results suggest that local effects make this inductive setting challenging, requiring carefully designed models and training procedures, but the performance of graph-based models remains competitive nonetheless.

\begin{table}
\caption{Performance comparison of deep learning models with ICON and persistence model baselines for temperature forecasting. Results report average $\pm$ 1 standard deviation computed over five runs with different random seeds, except for ICON and Chronos-2. For all seeds, the division seen-unseen stations is the same. The first part of the table reports the MAE, while the second part the Energy Score.}
\label{tab:temp}
\centering
\begin{tabular}{lllllll}
\toprule
& \multicolumn{2}{c}{All stations} & \multicolumn{2}{c}{Meteo stations} & \multicolumn{2}{c}{Rain gauges} \\
Model & \multicolumn{1}{c}{Seen} & \multicolumn{1}{c}{Unseen}  & \multicolumn{1}{c}{Seen} & \multicolumn{1}{c}{Unseen}  & \multicolumn{1}{c}{Seen} & \multicolumn{1}{c}{Unseen} \\
\midrule
\multicolumn{7}{c}{Mean Absolute Error (MAE)}\\
\midrule
PM-day      & 2.343\textsubscript{$\pm$0.000} & 2.410\textsubscript{$\pm$0.000} & 2.344\textsubscript{$\pm$0.000} & 2.451\textsubscript{$\pm$0.000} & 2.337\textsubscript{$\pm$0.000} & 2.368\textsubscript{$\pm$0.000} \\
PM-st       & 3.155\textsubscript{$\pm$0.000} & 3.045\textsubscript{$\pm$0.000} & 3.132\textsubscript{$\pm$0.000} & 3.106\textsubscript{$\pm$0.000} & 3.316\textsubscript{$\pm$0.000} & 2.983\textsubscript{$\pm$0.000} \\
Chronos-2   & 1.626 & 1.692 & 1.616& 1.679 & 1.700& 1.707\\
ICON        & 1.481 & 1.453 & 1.416\ & 1.371\ & 1.945 & 1.536 \\
\midrule
TCN         & 1.609\textsubscript{$\pm$0.002} & 1.762\textsubscript{$\pm$0.011} & 1.607\textsubscript{$\pm$0.002} & 1.722\textsubscript{$\pm$0.005} & 1.624\textsubscript{$\pm$0.003} & 1.803\textsubscript{$\pm$0.018} \\
MP-TCN      & 1.521\textsubscript{$\pm$0.007} & 1.726\textsubscript{$\pm$0.019} & 1.515\textsubscript{$\pm$0.007} & 1.718\textsubscript{$\pm$0.017} & 1.563\textsubscript{$\pm$0.007} & 1.734\textsubscript{$\pm$0.049} \\
DCRNN       & 1.488\textsubscript{$\pm$0.004} & 1.633\textsubscript{$\pm$0.007} & 1.483\textsubscript{$\pm$0.004} & 1.612\textsubscript{$\pm$0.009} & 1.523\textsubscript{$\pm$0.013} & 1.654\textsubscript{$\pm$0.019} \\
\toprule
\multicolumn{7}{c}{Energy Score (ES)}\\
\toprule
PM-day       & 1.772\textsubscript{$\pm$0.000} & 1.818\textsubscript{$\pm$0.000} & 1.771\textsubscript{$\pm$0.000} & 1.857\textsubscript{$\pm$0.000} & 1.776\textsubscript{$\pm$0.000} & 1.779\textsubscript{$\pm$0.000} \\
PM-st        & 2.363\textsubscript{$\pm$0.000} & 2.288\textsubscript{$\pm$0.000} & 2.346\textsubscript{$\pm$0.000} & 2.336\textsubscript{$\pm$0.000} & 2.482\textsubscript{$\pm$0.000} & 2.239\textsubscript{$\pm$0.000} \\
ICON         & 1.218\textsubscript{$\pm$0.000} & 1.200\textsubscript{$\pm$0.000} & 1.154\textsubscript{$\pm$0.000} & 1.115\textsubscript{$\pm$0.000} & 1.674\textsubscript{$\pm$0.000} & 1.287\textsubscript{$\pm$0.000} \\
\midrule
TCN          & 1.149\textsubscript{$\pm$0.002} & 1.266\textsubscript{$\pm$0.006} & 1.147\textsubscript{$\pm$0.002} & 1.231\textsubscript{$\pm$0.003} & 1.163\textsubscript{$\pm$0.003} & 1.301\textsubscript{$\pm$0.011} \\
MP-TCN       & 1.086\textsubscript{$\pm$0.005} & 1.248\textsubscript{$\pm$0.014} & 1.081\textsubscript{$\pm$0.005} & 1.237\textsubscript{$\pm$0.012} & 1.118\textsubscript{$\pm$0.006} & 1.258\textsubscript{$\pm$0.037} \\
DCRNN        & 1.062\textsubscript{$\pm$0.003} & 1.175\textsubscript{$\pm$0.009} & 1.059\textsubscript{$\pm$0.003} & 1.157\textsubscript{$\pm$0.008} & 1.087\textsubscript{$\pm$0.008} & 1.193\textsubscript{$\pm$0.021} \\
\bottomrule
\end{tabular}
\end{table}

\section{Illustrative forecast samples}\label{app:fcast-samples}

\begin{figure}[hb]
  \centering
  \begin{subfigure}[t]{0.98\textwidth}
      \centering
      \includegraphics[width=\textwidth]{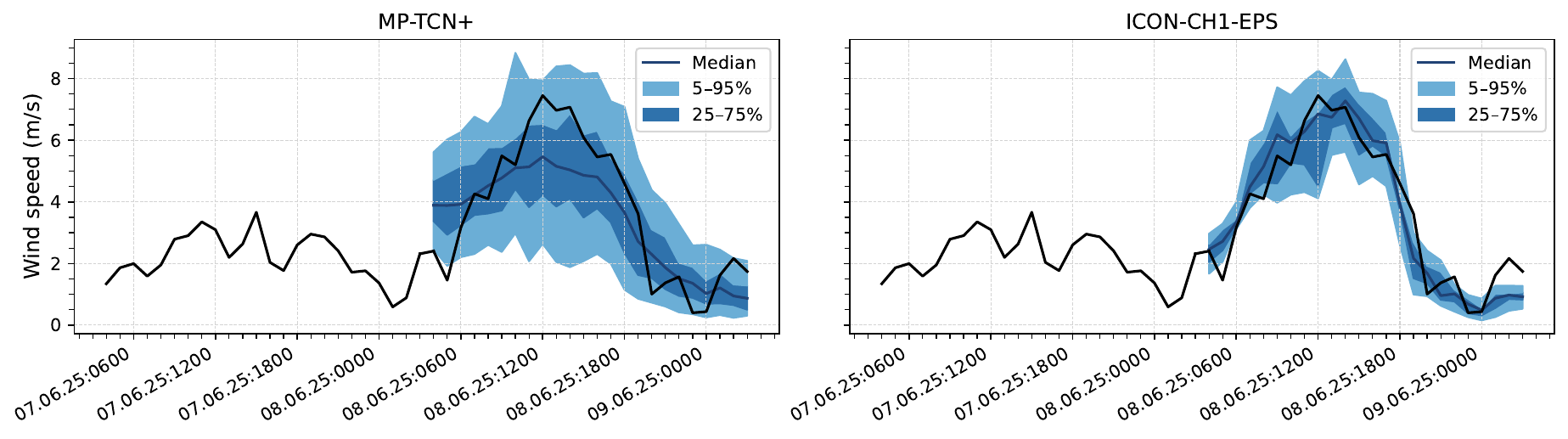}
  \end{subfigure}
\centering
  \begin{subfigure}[t]{0.98\textwidth}
      \centering
      \includegraphics[width=\textwidth]{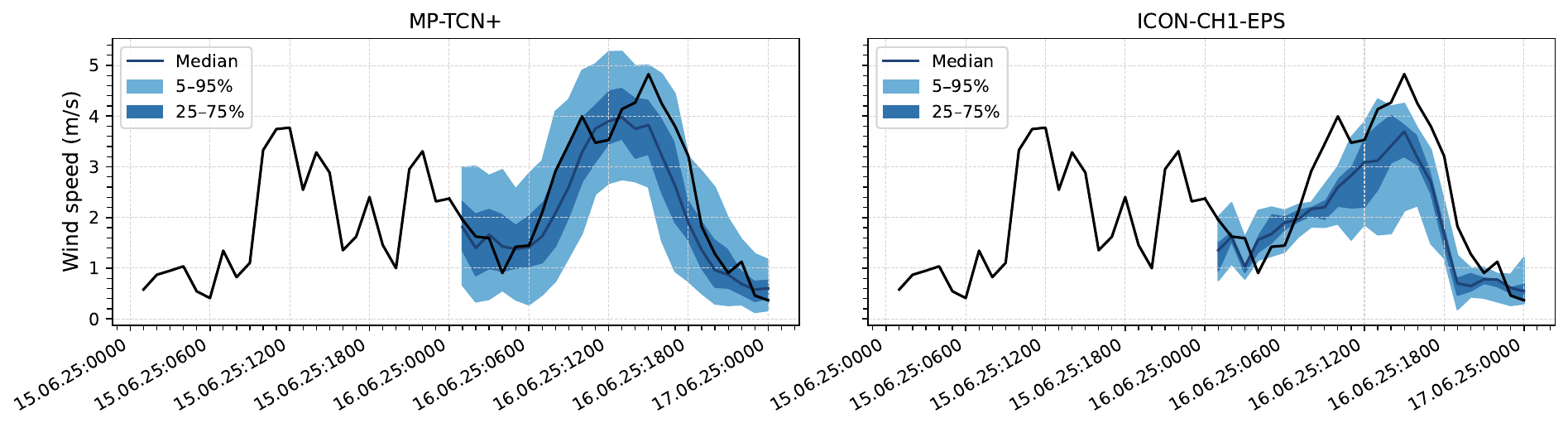}
  \end{subfigure}
\centering
  \begin{subfigure}[t]{0.98\textwidth}
      \centering
      \includegraphics[width=\textwidth]{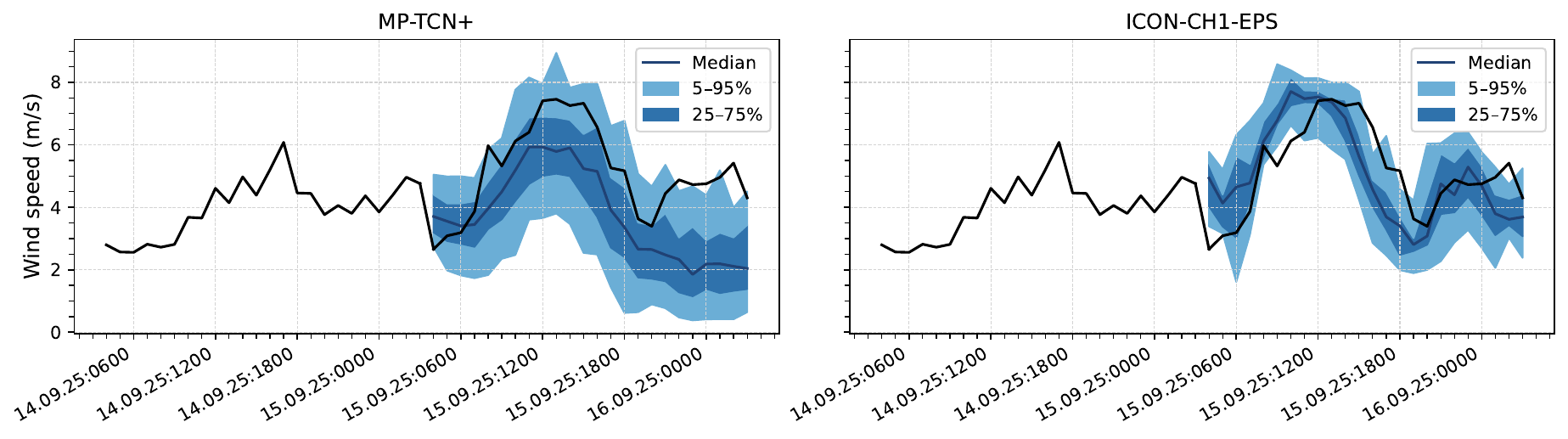}
  \end{subfigure}
  \centering
  \begin{subfigure}[t]{0.98\textwidth}
      \centering
      \includegraphics[width=\textwidth]{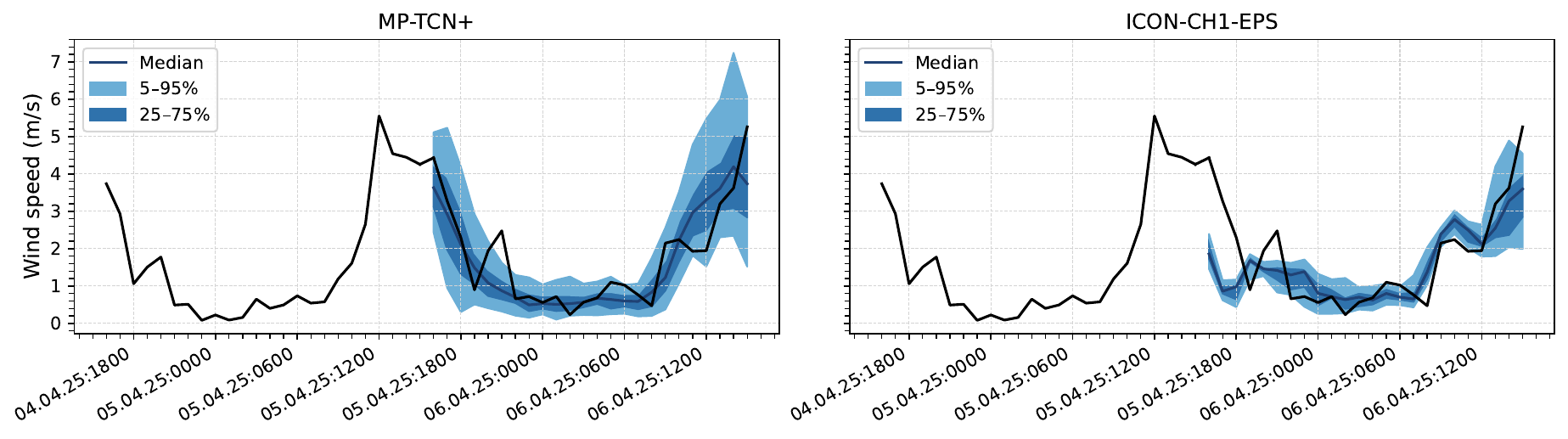}
  \end{subfigure}

\caption{Visual comparison between MP-TCN+ (left column) and ICON-CH1-EPS (right column) for the wind forecasting task on the test set aligned with the ICON model. Wind speed is obtained as $\sqrt{u^2 + v^2}$ from the eastward $u$ and northward $v$ wind components. For MP-TCN+, 100 predictive samples were produced, whereas ICON-CH1-EPS uses its 11 ensemble members. Shaded areas denote forecast uncertainty, in particular the 5th, 25th, 75th and 95th quantiles, as well as the median member in a solid line. The observed wind speed is displayed with a solid black line.}
  \label{fig:forecasts}
\end{figure}

\begin{figure}[hb]
  \centering
  \begin{subfigure}[t]{0.98\textwidth}
      \centering
      \includegraphics[width=\textwidth]{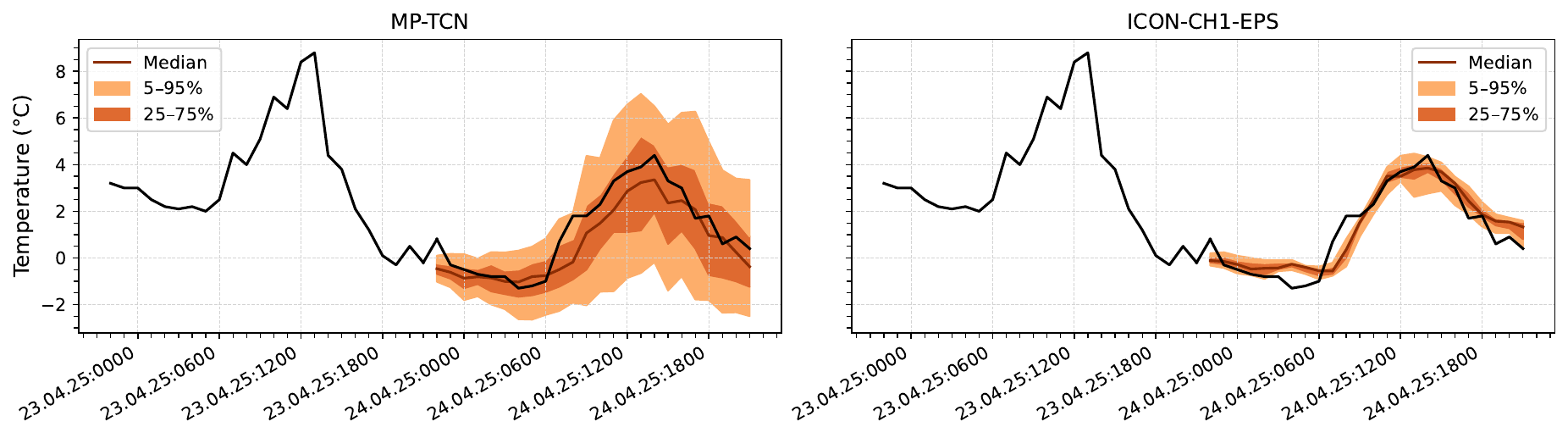}
  \end{subfigure}
\centering
  \begin{subfigure}[t]{0.98\textwidth}
      \centering
      \includegraphics[width=\textwidth]{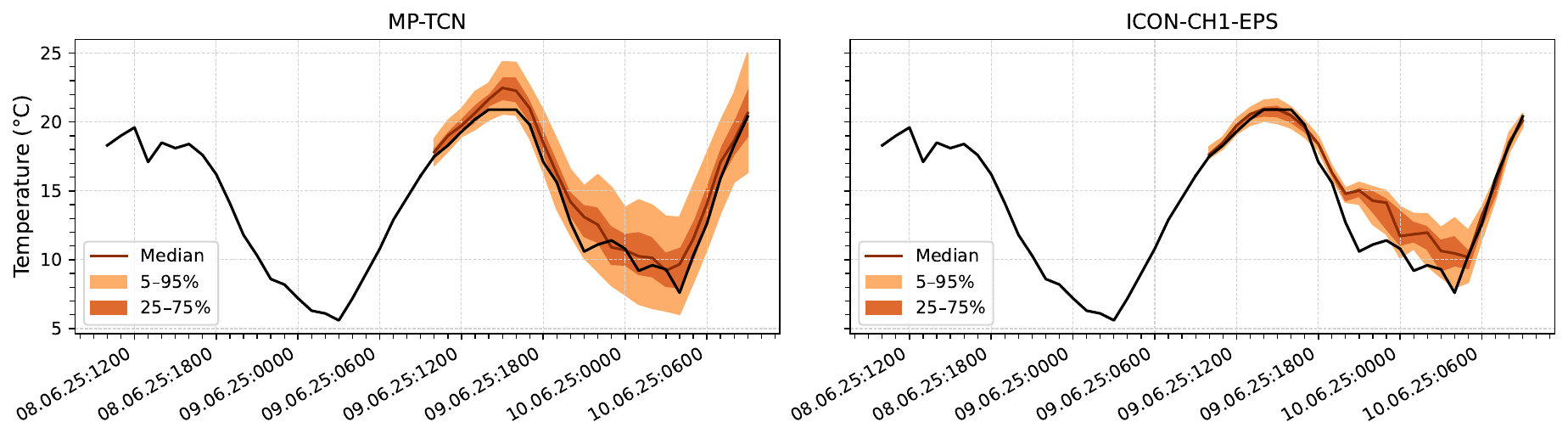}
  \end{subfigure}
\centering
  \begin{subfigure}[t]{0.98\textwidth}
      \centering
      \includegraphics[width=\textwidth]{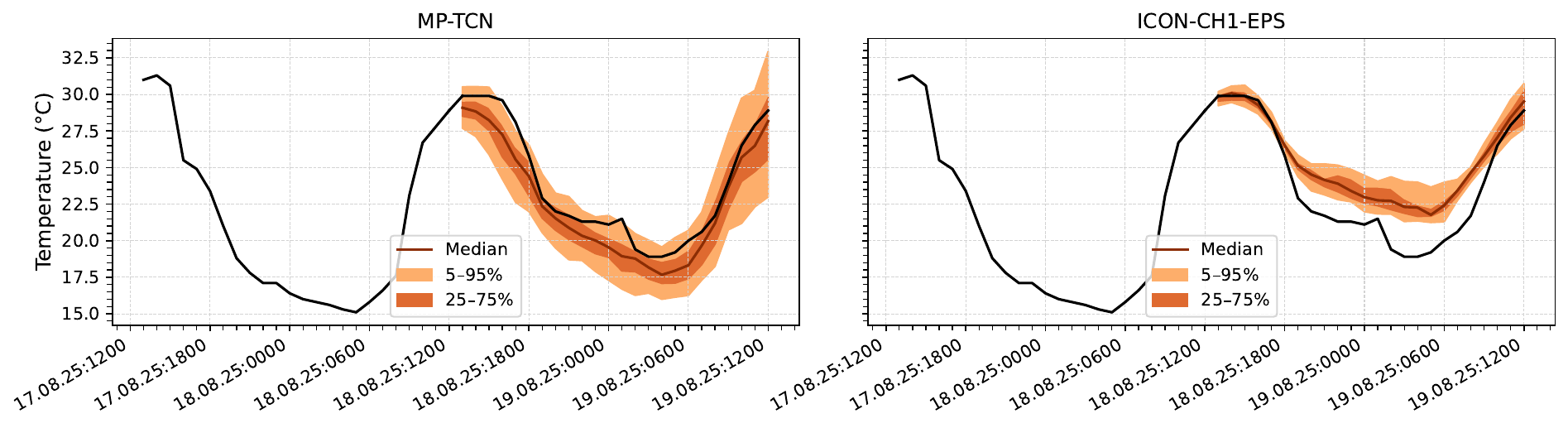}
  \end{subfigure}
\centering
  \begin{subfigure}[t]{0.98\textwidth}
      \centering
      \includegraphics[width=\textwidth]{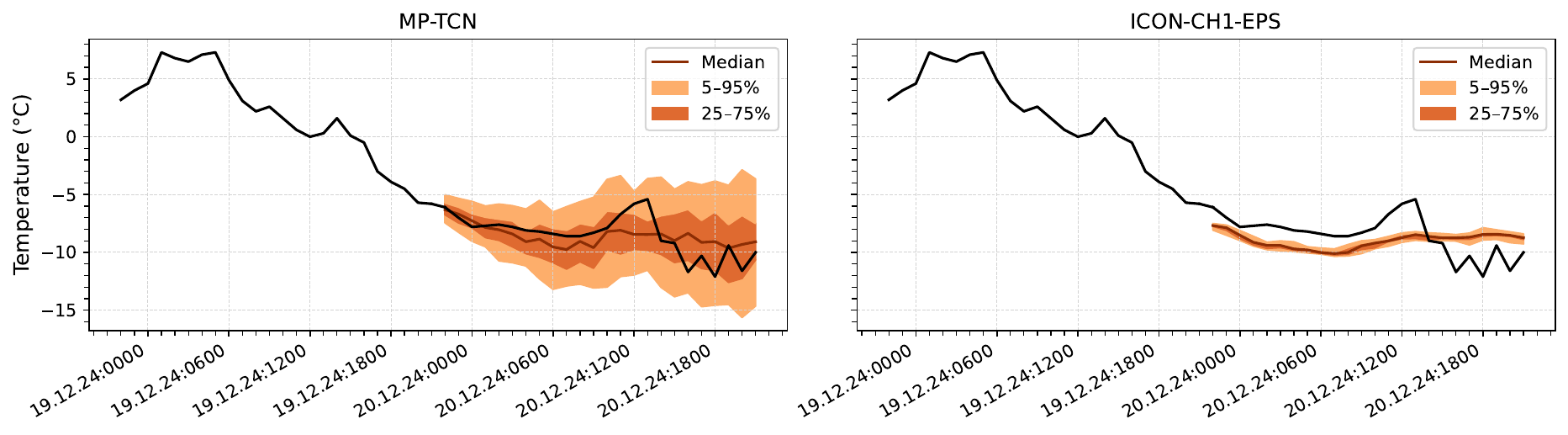}
  \end{subfigure}
\centering

\caption{Visual comparison between MP-TCN (left column) and ICON-CH1-EPS (right column) for the inductive learning task of temperature forecasting. Results are shown for a subset of meteorological stations and rain gauges that were held out from the training dataset; MP-TCN was therefore not trained on observations from these locations.}
  \label{fig:forecasts_temp}
\end{figure}

Figure~\ref{fig:forecasts} displays wind speed forecasts for 24h derived from the eastward and northward wind components for some sampled forecasting windows in the test set at different station locations. These forecasts come from the experiment outlined in Section \ref{sec:experiments}. The figure compares MP-TCN+ against the physics-based ICON model. Forecast uncertainty is also visualized as shaded areas, representing the the 5th, 25th, 75th and 95th quantiles computed from the samples.
Similarly, Figure~\ref{fig:forecasts_temp} displays the results of MP-TCN and ICON on the inductive learning task for temperature forecasting (Appendix \ref{app:temperature-extra}). The figure shows predictions at meteorological stations and rain gauges that were not part of the training set, illustrating the model's ability to generalize to previously unseen locations.

\vspace{1cm}

\end{document}